\documentclass[conference]{IEEEtran}
\IEEEoverridecommandlockouts
\usepackage{cite}
\usepackage{amsmath,amssymb,amsfonts}
\usepackage{algorithmic}
\usepackage{graphicx}
\usepackage{textcomp}
\usepackage{xcolor}

\usepackage{latexsym}

\usepackage{comment}
\usepackage{amsmath}
\usepackage{ stmaryrd }

\usepackage{amssymb}
\usepackage{dsfont}
\usepackage{subfigure}
\usepackage{multirow}
\usepackage{bm}

\usepackage{cite}
\usepackage{hyperref}
\def\BibTeX{{\rm B\kern-.05em{\sc i\kern-.025em b}\kern-.08em
    T\kern-.1667em\lower.7ex\hbox{E}\kern-.125emX}}
\begin{document}

\title{Causal Fairness-Guided Dataset Reweighting using Neural Networks
}

\author{\IEEEauthorblockN{Xuan Zhao}
\IEEEauthorblockA{
\textit{SCHUFA Holding AG}\\
Germany \\
xuan.zhao@schufa.de}
\and
\IEEEauthorblockN{Klaus Broelemann}
\IEEEauthorblockA{
\textit{SCHUFA Holding AG}\\
Germany \\
Klaus.Broelemann@schufa.de}
\and
\IEEEauthorblockN{Salvatore Ruggieri}
\IEEEauthorblockA{
\textit{University of Pisa}\\
Italy \\
salvatore.ruggieri@unipi.it}
\and
\IEEEauthorblockN{Gjergji Kasneci}
\IEEEauthorblockA{
\textit{Technical University of Munich}\\
Germany \\
gjergji.kasneci@tum.de}
}

\maketitle

\begin{abstract}
The importance of achieving fairness in machine learning models cannot be overstated. Recent research has pointed out that fairness should be examined from a causal perspective, and several fairness notions based on the on Pearl's causal framework have been proposed. In this paper, we construct a reweighting scheme of datasets to address causal fairness. Our approach aims at mitigating bias by considering the causal relationships among variables and incorporating them into the reweighting process. 
%which can manipulate the data distribution of a given causal graph by reweighting of the original data, while also ensuring various causal fairness definitions. 
The proposed method adopts two neural networks, whose structures are intentionally used to reflect the structures of a causal graph and of an interventional graph. The two neural networks can approximate the causal model of the data, and the causal model of interventions. Furthermore, reweighting guided by a  discriminator is applied to achieve various fairness notions. Experiments on real-world datasets show that our method can achieve causal fairness on the data while remaining close to the original data for downstream tasks. 
\end{abstract}.


\section{Introduction} \label{sec:intro}

%1. generative vs reweighting computational expense

%Fairness in machine learning is receiving significant attention in the literature.
Pre-processing data to satisfy fairness requirements is an important research question in machine learning. Models trained on biased data may learn such biases and generalize them, thus leading to discriminatory decisions against socially sensitive groups defined on the grounds of gender, race and age, or other protected grounds \cite{pedreshi2008,zliobaite2011,hardt2016a,zhang2016,zhang2018}.
Many methods  have been proposed to modify the training data in order to mitigate biases and to achieve specific fairness requirements \cite{feldman2015,zhang2016,zhang2019a,edwards2016,xie2018,madras2018c,zhang2018a}.

%These methods include: Massaging \cite{kamiran2009}, Reweighting \cite{calders2009}, Sampling \cite{kamiran2012a}, Disparate Impact Removal \cite{feldman2015}, Causal-based Removal \cite{zhang2016,zhang2019a} and Fair Representation Learning \cite{edwards2016,xie2018,madras2018c,zhang2018a}.

For reliable and effective treatment, particularly in a legal context, discrimination claims usually require demonstrating causal relationships between sensitive attributes and questionable decisions (or predictions), instead of mere associations or correlations. Compared with the fairness notions based on correlation, causality-based fairness notions and methods include additional knowledge of the causal structure of the problem. This knowledge often reveals the mechanism of data generation, which helps comprehend and interpret the influence of sensitive attributes on the output of a decision process. Causal fairness seeks to address the root causes of disparities rather than simply trying to eliminate them in a post-hoc manner. 

We draw upon the ideas and concepts presented in CFGAN~\cite{xu2019b} as the framework for our research. Instead of fair dataset generation in CFGAN, however,  we propose a method which reweighs the samples to achieve fairness criteria with the help of two neural networks to reflect the causal and interventional graphs, and a discriminator to guide the reweighting. As the general requirement of modifying datasets is to preserve the data utility as much as possible for the downstream tasks. The intuition of the reweighting scheme is that in a given dataset, there are individuals who are treated `fairer' in the causal mechanism and by assigning higher weights to these individuals, we could slightly alter the underlying causal mechanism to achieve fairness and do not influence much on the performance of downstream tasks. In this case, hopefully we could mitigate the historical bias. In addition, by analyzing the high/low weights assigned to samples, a reweighting method  like ours enables for a high-level understanding the biases. 

The experiments (Section \ref{experiment}) show that reweighed data outperform generated data in utility. 
%
%In the taxonomy of preprocessing, in-processing and post-processing methods \cite{roh2021a,calmon2017a,aghaei2019a,berk2017a,hardt2016b} for bias mitigation, our method falls into the category of preprocessing since the weights of the samples are kept for further down-stream task. 
%
In the taxonomy of pre-processing, in-processing and post-processing methods for bias mitigation \cite{roh2021a,calmon2017a,aghaei2019a,berk2017a,hardt2016a}, our method falls into the category of pre-processing, as we deal with the dataset before it is given in input to the downstream learning algorithm. Thus, our approach is model-agnostic, as any pre-processing method. 

We summarize our contribution as follows: (1) We formulate a novel and sample-based reweighting
method for mitigating different causal bias related to sensitive groups. (2) We show that by simulating the underlying causal model
that reflects the causal relations of the real data, and the causal
model after the intervention, with the help of a discriminator, our reweighting approach leads to fair reweighted data. (3) We provide a thorough evaluation of the proposed technique on benchmark datasets and show the viability of our approach.

%The remainder of this paper is structured as follows. In Section \ref{background},
%we introduce the relevant foundation of our work. In Section \ref{method}, we formalize our adversarial reweighing approach. In
%Section \ref{experiment}, we present a detailed analysis of the performance of the method on two tabular datasets. Finally,
%Section \ref{conclusion} provides discussion and conclusion of the paper.

\section{Preliminary}\label{background}

Throughout this paper, we consider a structural causal model $\mathcal{M} =
\langle U, V, F \rangle$, that is learned from a dataset $\mathcal{D}=\{(s_k,x_k,y_k)\}_{k=1}^m$ where $s_k\in S=\{0,1\},x_k\in X \subseteq\mathbb{R}^d, y_k\in Y=\{0,1\}$.

1) $U$ denotes exogenous variables that cannot be observed but constitute the background knowledge behind the model. $P (U)$ is a joint probability distribution of the variables in $U$.

2) $V$ denotes endogenous variables that can be observed. %denotes the observed random variables determined by variables in $U \cup V$. 
 In our work, we set $V = \{S, X, Y\}$. $S$ represents the sensitive attribute, $Y$ represents the outcome attribute, and $X$ represents all other attributes. Additionally, $s^+$ is used to denote $S = 1$ and $s^-$ to denote $S = 0$.

3) $F$ denotes the deterministic functions. For each $V_i \in V$, there is a corresponding function $f_{V_i}$ that maps from domains of the variables in $Pa_{V_i} \cup U_{V_i}$ %$U\cup(V\backslash\{V_i\})$
to $V_i$, namely $V_i = f_{V_i} (Pa_{V_i}, U_{V_i})$. Here, $Pa_{V_i} \subseteq V\backslash V_i$ represents the parents of $V_i$, and $U_{V_i}$ also represents the parents (exogenous variables) of $V_i$,  $U_{V_i} \subseteq U$.

We denote by $\mathcal{G}$ the causal graph $\mathcal{G}$ associate with $\mathcal{M}$, and assume it is a Directed Acyclic Graph (DAG).

\subsection{Causal Fairness Criteria}\label{causal_notation}

To understand causal effects in the causal model $\mathcal{M}$, we can use the do-operator \cite{pearl2009}, which represents a physical intervention that sets a variable $S \in V$ to a constant value $s$. By performing an intervention $do(S = s)$, we replace the original function $S = f_S (Pa_{S} , U_S)$
with $S = s$. This results in a change in the distribution of all variables that are descendants of $S$ in the causal graph. $\mathcal{M}_s$ is the interventional causal model and its corresponding graph $\mathcal{G}_s$ the interventional graph. In $\mathcal{G}_s$, edges to $S$
are deleted according to the definition of intervention and $S$ is replaced with constant $s$. The interventional distribution for $Y$ is denoted by
$P (Y|do(S = s))$. Using the do-operator, we can compare the interventional distributions under different interventions to infer the causal effect of $S$ on $Y$. In this paper, we focus on the following causal causal fairness notions:

\paragraph{Total effect} 
The total effect infers the causal effect of $S$ on $Y$ through all possible causal paths from $S$ to $Y$. 
The total effect of the difference of $s^-$ to $s^+$ on $Y$ is given by $TE(s^+, s^-) = P(Y_{s^+} ) -P(Y_{s^-})$, where $P(\cdot)$ here refers to the interventional distribution probability. Total fairness is satisfied if $|TE(s^+, s^-)|<\tau$ ($\tau$ is the fairness threshold). Note that statistical parity is similar to total effect but is fundamentally different. Statistical parity measures the conditional distributions of $Y$ change of the sensitive attribute
from $s^-$ to $s^+$.

\paragraph{Path-specific fairness} The path-specific effect is a fine-grained assessment of causal effects, that is, it can evaluate the causal effect transmitted along certain paths. It is used to distinguish among direct discrimination, indirect discrimination, and explainable bias. It infers the causal effect of $S$ on $Y$ through a subset of causal paths from $S$ to $Y$, which is referred to as the $\pi$-specific effect denoting the subset of causal paths as $\pi$. The specific effect of a path set $\pi$ on $Y$, caused by changing the value of $S$ from $s^-$ to $s^+$ with reference to $s^-$, is given by the difference of the interventional distributions: $SE_\pi (s^+, s^-) = P (Y_{s^+|\pi, s^-|\overline{\pi}} ) - P (Y_{s^-} )$, where $P (Y_{s^+|\pi, s^-|\overline{\pi}} )$ represents the distribution resulting from intervening $do(s^+)$ only along the paths in $\pi$ while $s-$ is used as a reference through other paths $\overline{\pi}$. If $\pi$ contains all direct edge from $S$ to $Y$, $SE_\pi (s^+, s^-)$ measures the direct discrimination. If $\pi$
contains all indirect paths from $S$ to $Y$ that pass through proxy attributes, $SE_\pi (s^+, s^-)$ evaluates the indirect discrimination. %If $\pi$ contains all indirect paths $S$ to $Y$ that pass through proxy attributes, $SE_\pi (s^+, s^-)$ assesses the explainable bias. 
Path-specific fairness is met if $|SE_\pi (s^+, s^-)|<\tau$.

\paragraph{Counterfactual fairness}
The counterfactual effect of changing $S$ from $s^-$ to $s^+$ on $Y$ under certain conditions $O=o$ (where $O$ is a subset of observed attributes $O\subseteq X$) for an individual with features $o$ is given by the difference between the interventional distributions $P(Y_{s^+}|o)$ and $P(Y_{s^-}|o)$: $CE(s^+, s^-|o) = P(Y_{s^+}
|o)-P(Y_{s^-}|o)$. Counterfactual fairness is met if $|CE(s^+, s^-|o)|<\tau$. Any context $O=o$
represents a certain sub-group of the population, specifically, when $O=X$, it represents specific individual(s).

\subsection{Causal Discovery}\label{discovery}

Methods for extracting a causal graph from given data (causal discovery) can be broadly categorized into two  constraint-based and score-based methods \cite{spirtes2016,glymour2019}. Constraint-based methods, such as \cite{spirtes2013,spirtes1991,colombo2012}, utilize conditional independence tests under specific assumptions to determine the Markov equivalence class of causal graphs. Score-based methods, like \cite{vowels2021}, evaluate candidate graphs using a pre-defined score function and search for the optimal graph within the space of DAGs. Such an approach is formulated as a combinatorial optimization problem:

\begin{align} \label{eq:search}
%\underset{\mathcal{G}}{\text{min}} & Score(\mathcal{G};V) =\mathcal{L}(\mathcal{G};V)+\lambda \mathcal{R}_{sparse}(\mathcal{G}),\nonumber \\
\min_\mathcal{G}\, & Score(\mathcal{G};V) =\mathcal{L}(\mathcal{G};V)+\lambda \mathcal{R}_{sparse}(\mathcal{G}),\nonumber \\
s.t.\ & \mathcal{G} \in DAG
\end{align}

In the realm of causal discovery, the problem can be divided into two components, which constrain the score function $Score(\mathcal{G}; V)$  and $\mathcal{G} \in DAG$. The score function is comprised of: (1) the goodness-of-fit $\mathcal{L}(\mathcal{G}; V) = \frac{1
}{m}
\sum_{k=1}^{m}
 l(v_k, F(v_k))$ is the loss of fitting observation of
$v_k$; $F$ denotes the deterministic functions as defined earlier in Section \ref{background} (2) the sparsity $\mathcal{R}_{sparse}(\mathcal{G})$ which regulates the number of edges in $\mathcal{G}$. $\lambda$ serves as a hyperparameter that controls the regularization strengths.

In this work, we assume that the given causal graph $\mathcal{G}$ is learned from a score-based causal discovery, so $\mathcal{G}$ should have goodness-of-fit and sparsity. 

\subsection{Intervention through Controlled Neural Networks}\label{neural}

In CausalGAN \cite{kocaoglu2017}, a noise vector $Z$ is partitioned into $\{Z_{V_1}, Z_{V_2}, ... , Z_{V_{|V|}}\}$ to mimic the exogenous variables $U$ in the structural causal model $\mathcal{M}$ described in Section~\ref{background}. The generator $G(Z)$ contains $|V|$ sub-neural networks $\{G_{V_1}, G_{V_2}, ... , G_{V_{|V|}}\}$ to generate the values of each node $V_i$ in the graph. The input of $G_{V_i}$ is the output of $G_{Pa_{V_i}}$ combined with $Z_{V_i}$. Here, $G_{V_i}$ is trying to approximate the corresponding function $f_{V_i} (Pa_{V_i}, U_{V_i})$ in the causal model $\mathcal{M}$.  %the  If there is a directed edge from node $V_j$ to $V_i$ in the causal graph, then the output of $G_{V_j}$ is used as an input to $G_{V_i}$ to reflect this causal relationship. 
The adversarial game is played to ensure that the generated observational distribution is not differentiable from the real observational distribution. %It indicates that this aligns with any causal model that portrays the identical causal graph structure, when the following conditions are met: (1) $P(V)$ is strictly positive; (2) The links between the sub-neural networks $G_{V_i}$ are arranged to mirror the causal graph structure; and (3) the generated observational distribution corresponds to the original observational distribution. 
In the work of CFGAN, two generators are used to simulate the causal model $\mathcal{M}$ and the interventional model $\mathcal{M}_s$, while two discriminators try to maintain that: (1) the generated data is close to the orginal distribution, and (2) the causal effect is mitigated. In our work, we also use a similar design but we do not model the noise $Z$ since our goal is not to generate fairness-aware data, but to reweigh the given data. %to simulate both the causal graph $\mathcal{G}$ and the interventional graph $\mathcal{G}_s$ with corresponding neural networks synchronized to maintain the connections between the two causal graphs.

\section{A Reweighting Approach for Different Causal Fairness Critiria}\label{method}

\subsection{Problem Formulation}\label{form}
As mentioned in Section \ref{background}, the notation used in our work is based on the conventional approach. %For $\{S,Y\}\in V$, $S$ denotes the sensitive variable, $Y$ denotes the decision variable. %and $\overline{V}$ denotes the set of all other variables. 
We are given a causal graph $\mathcal{G}$ and a dataset $\mathcal{D}$ with $m$ i.i.d. samples drawn from $P(V)$. We assume that $\mathcal{G}$ is sufficient to describe the causal relationships between the variables $V$. In this paper, we build our method on a causal graph of observational data, so we do not specifically model $U$. The problem we are facing is that from the given causal graph $\mathcal{G}$, $S$ has a causal effect on $Y$. Our method aims to achieve two objectives: (1) preserve the goodness-of-fit (mentioned in Section \ref{discovery}) by maintaining the empirical reweighted data distribution close to the  original data distribution for utility of the downstream tasks; and (2) ensure that $S$ cannot be used to discriminate when predicting $Y$ based on various causal criteria in the interventional model $\mathcal{M}_s$. We treat $S$ and $Y$ as binary variables in this paper. 
However, this can be easily extended to multi-categorical or numerical cases. 
Also, we focus on the causal effect of $S$ on $Y$, but the model can deal with causal effects among multiple variables. We try to reach the following causal fairness notions mentioned in Section \ref{causal_notation}, including total fairness \cite{zhang2018b}, path-specific fairness (elimination of indirect discrimination) \cite{zhang2016}, and counterfactual fairness \cite{kusner2018}.

\subsection{Reweighting For Causal Fairness}

\begin{figure}
    \centering
    \includegraphics[width=.47\textwidth]{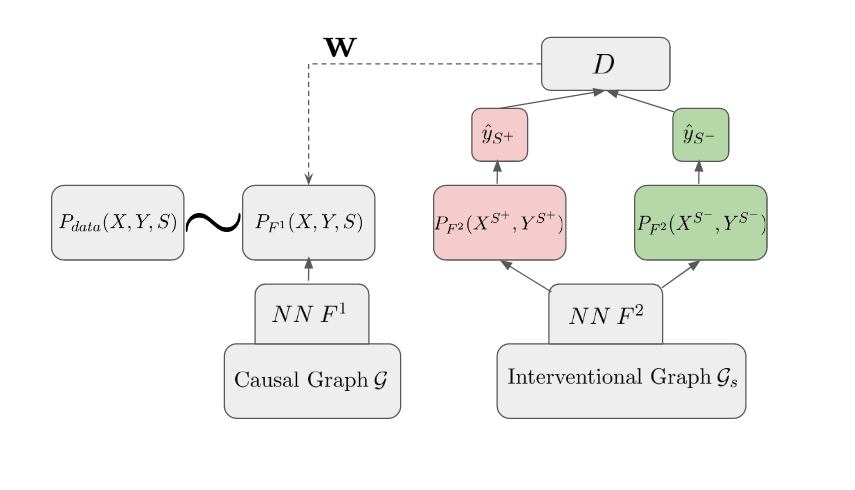}
    \caption{The framework of reweighting: the structure of NN (neural network) $F^1$ reflects the original causal graph $\mathcal{G}$; the structure of NN $F^2$ refelects the interventional causal graph $\mathcal{G}_s$; the discriminator $D$ tells if a $\hat{y}$ estimated by $F^2$ is from the group $S^+$ or the group $S^-$. An adverserial game is played between the reweighting on the data samples and $D$ to reach a situation where $D$ is not capable of differentiating whether $y$ is from $S^+$ or $S^-$ and a specific causal fairness is reached. The weights of samples are also forwarded to $F^1$ to make sure that the reweighted empirical data distribution is close to the original data distribution from which the causal graph $\mathcal{G}$ is learned.}
    \label{fig:flow}
\end{figure}

We propose a reweighting scheme which consists of neural networks ($F^1$, $F^2$) and one discriminator ($D$). Fig. \ref{fig:flow} shows the framework of our method.

As shown in Section \ref{form}, causal fairness notions measures the difference between the interventional distributions. To guarantee these notions, our method adopts two neural networks to approximate the causal relations. One neural network $F^1$ simulates the causal model $\mathcal{M}$, while the other neural network $F^2$ approximates the interventional model $\mathcal{M}_s$ according to which kind of causal effect is measured. $F^1$ aims to force the reweighted data close to the given causal graph, and $F^2$ aims to drive the interventional distributions to satisfy the specific notion defined in Section
\ref{form}. To represent the connections between the two causal models, the two neural networks share certain structures and parameters, while they differ in sub-neural networks to indicate the intervention (the edges to $S$ in the interventional graph is deleted). Then, our method adopts a discriminator $D$ trying to distinguish the two interventional distributions (reweighted) $P(Y_{s^+})$ and $P(Y_{s^-})$. Finally, the discriminator and reweighting play an adversarial game to produce weights for individuals in the dataset.

\begin{figure}
\centering     %%% not \center
\subfigure[Causal Graphs $\mathcal{G}$ and $\mathcal{G}_s$]{\label{fig:total_a}\includegraphics[width=50mm,height=1.2in]{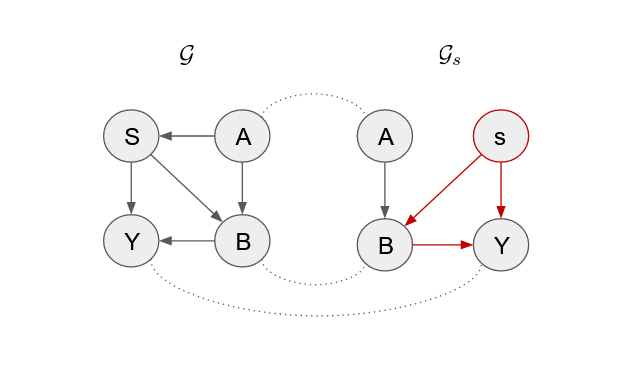}}
\subfigure[Neural Networks $F^1$ and $F^2$]{\label{fig:total_b}\includegraphics[width=70mm,height=1.5in]{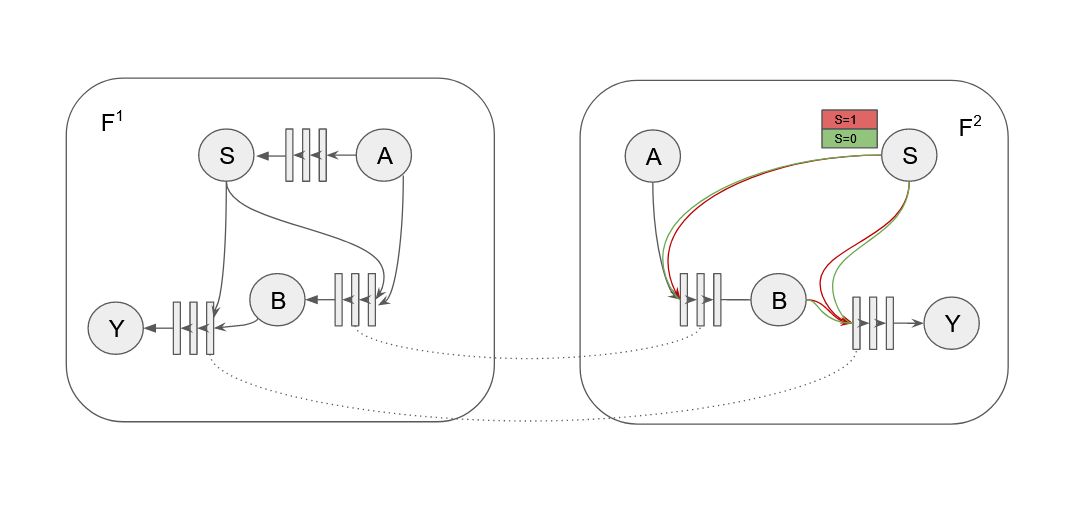}}
\caption{The Neural Networks $F^1$
and $F^2$ on total effect. $S$ is 1 or 0 for the interventional joint distributions $P_{F^2} (s^+)$ (red path) and $P_{F^2} (s^- )$  (green path), respectively. The pair of nodes connected by dashed lines indicate that they share the same function (structures and parameters of the corresponding sub-neural networks).
} \label{fig:graph2}
\end{figure}

To better illustrate our design, we divide $X$ into $\{A, B\}$ and  $V=\{S,A, B, Y\}$ based on the positions of the nodes in the causal graph -- variables in $A$ are direct causes of $S$ and variables in $B$ are descendants of $S$ and $A$.

\begin{comment}

Consider an example in Figure \ref{fig:graph2} which involves 4 variables
$\{A, S, B, Y \}$. Figure \ref{fig:graph2} shows the causal graph $\mathcal{G}$ (we assume that it is causally sufficient) and the interventional graph $\mathcal{G}_s$ under $do(S = s)$, where the double headed arrows indicate the pair of nodes that share the same function (structures and parameters of the corresponding neural networks). As shown, the
edge from $A$ to $S$ is deleted in $\mathcal{G}_s$, which is also reflected in $F^2$. In addition, for each pair of nodes in the graphs, e.g., $B$
in $\mathcal{G}$ and $B$ in $\mathcal{G}_s$, the corresponding sub-neural networks are also synchronized during the training process, e.g., $F^1_B$ and $F^2_B$.
\end{comment}

%\textbf{Total Effect}
%Total effect that encodes a forbidden edge between two variables $S, A \in V$ implies that neither $S$ nor $A$ can be a member of each other’s parent set.

%We use (sequential) neural networks $f$ to approximate $\mathcal{G}$ and in $\mathcal{G}_s$, the corresponding components of neural networks are kept the same except for the sub-neural network mimicking the edge from A to s. In addition, we use a discriminator in $\mathcal{G}_s$ to make sure that the $y$ is not differentiable whether if $s=1$ or $s=0$.

\subsubsection{Reweighting for Total Fairness}\label{sec:total}

The causal graph $\mathcal{G}$ is shown in Fig. \ref{fig:total_a}.  We also show the interventional graph $\mathcal{G}_s$ with the intervention $do(S=s)$ and the edge from $A$ to $S$ is deleted in $\mathcal{G}_s$, which is also altered in $F^2$. The pair of nodes connected by dashed lines indicate that they share the same function (structures and parameters of the corresponding sub-neural networks) as shown in Fig. \ref{fig:total_b}. For parallel nodes in the two graphs, the corresponding sub-neural networks are synchronized during the training process.

We first show our method to achieve total fairness by describing each components of our design. As mentioned in Section \ref{causal_notation} , $|TE(s^+, s^-)|<\tau$ must hold for all possible paths from $S$ to $Y$ shown in Fig. \ref{fig:total_a}.

\paragraph{Neural Networks $F^1$ and $F^2$} 

The feed-forward Neural Network $F^1$ is constructed to correspond with the causal graph $\mathcal{G}$. It consists of $|V|-r$ sub-neural networks ($r$ is the total number of the root nodes in $\mathcal{G}$), with each corresponding to a node in $V$ (expect for the root nodes). Similar to what is described as the design of CFGAN in Section \ref{neural}, each sub-neural network $F^1_{V_i}$ is trying to approximate the corresponding function $f_{V_i} (Pa_{V_i})$ in the causal model $\mathcal{M}$ of the given causal graph $\mathcal{G}$. When $F^1$ is properly trained, the causal model $\mathcal{M}$ is learned. Then, $F^1_{V_i}$ outputs the estimated values of $V_i$, i.e., $\hat{v_i}$. The other neural network $F^2$ is constructed to align with the interventional graph $\mathcal{G}_s$, where all the incoming edges to S are removed under the intervention $do(S=s)$. The layout of $F^2$ is analogous to $F^1$, but with the exception that the sub-neural network $F^2_S$ is designated as $F^2_S \equiv 1$ if $s=s^+$, and $F^2_S \equiv 0$ if $s=s^-$. To synchronize the two neural networks $F_1$ and $F_2$, they share the identical set of structures and parameters for every corresponding pair of sub-neural networks, i.e., $F^1_{V_i}$ and $F^2_{V_i}$ for each $V_i$ except for $S$. When $F_2$ is properly trained, the interventional model $\mathcal{M}_s$ is learned. With $\mathcal{M}$ and $\mathcal{M}_s$ learned, we could manipulate the interventional distributions to reach our goal of causal fairness. %As a result, $F^1$ can approximate the causal graph of the observational distribution, and $F^2$ can approximate the intervention graph of two interventional distributions, i.e., $(s, \hat{x}, \hat{y}) \sim P_{F^1} (S, X, Y)$,
%$(\hat{x}^{s^+} , \hat{y}^{s^+}  ) \sim P_{F^2} (x^{s^+} , y^{s^+} )$, if $s = s^+$, $(\hat{x}^{s^-} , \hat{y}^{s^-} ) \sim
%P_{F^2} (X_{s^-} , Y_{s^-})$, if $s = s^-$.

\paragraph{Discriminator}

 $D$ is used to differentiate between the two interventional distributions $\hat{y}_{s^+} \sim P_{F^2}(Y_{s^+})$ and $\hat{y}_{s^-} \sim P_{F^2}(Y_{s^-})$. The aim of the discriminator $D$ to minimize the bias by penalizing differences between both groups.

\paragraph{Weights}

Assuming the to-reach-causal-fairness-importance of each individual in the given dataset is known, we can assign importance to different individuals in $\mathcal{M}_s$ to improve causal fairness for any downstream task. $\mathbf{w} = (w_1, ... , w_m)$ is a sample reweighting vector with length $m$, where $w_k$ indicates the importance of the $k$-th observed sample $(s_k,x_k,y_k)$. We want to reach a balance of goodness-of-fit to the known causal graph $\mathcal{G}$ which is learned from $\mathcal{D}$ and reweighting for causal fairness. 

\textit{Recall that here we assume that the known causal graph $\mathcal{G}$ is learned from a causal discovery which means it achieves goodness-of-fit.} We do not want the reweighted data to drift too far from the original causal graph. We use hatted variables to represent the output of the neural networks of the graphs. To reach this objective, we have:

\begin{equation}\label{eq:net}
    S_{F^1}(\mathcal{G})=\underset{F^1}{\text{min}} \sum_{i=1}^{m}w_il((s_i,x_i,y_i),(s_i,\hat{x_i},\hat{y_i}))
\end{equation}

 where $l((s_i,x_i,y_i),(s_i,\hat{x_i},\hat{y_i}))$ represents the loss of fitting observation
$(x_i,y_i,s_i)$. In the experiment, we use weighted MSE loss for the continuous variables and weighted cross entropy loss for the categorical variables.
The problem then becomes how to learn appropriate the sample reweighting vector $\mathbf{w}$ for the objective of causal fairness. %$\lambda$ is a hyperparameter which controls a trade-off between utility and fairness. 
We formulate our objective as a minmax problem to reweight with $\mathcal{M}_s$:

\begin{equation}\label{eq:dis}
    \underset{\mathbf{w}}{\text{min}} \underset{D}{\text{max}}\sum_{k=1}^{m}w_k(D(\hat{y}^{s^+}_k)-D(\hat{y}^{s^-}_k)),
\end{equation}

To avoid information loss by assigning close to zero weights to some samples from the group of $S^+$, we introduce a regularization constraint to the minimization term: 

\begin{equation}\label{eq:balance}
    \sum_{k=1}^{m}(w_k-1)^2\leqslant Tm
\end{equation}

Thus, by adjusting the value of $T$, we can balance between similarity
and dissimilarity of the weights of samples.

Samples easily fitted with fairness constraint should contribute more to $\mathcal{G}_s$: these are the samples with less difference of discriminator outputs from $do(S=s^-)$ to $do(S=s^+)$. We therefore use downweighting on the not-hold-fairness samples, and upweighting on the hold-fairness samples. This could be achieved by assigning weights to samples based on the discriminator $D$ outputs. When the neural networks are properly trained, the discriminator should not be able to tell if the sample is from the group of $S^+$ or $S^-$ which could achieve total fairness as we describe in Section \ref{form}.

\subsubsection{Reweighting for Path-Specific Fairness}\label{sec:indirect}

The notions of direct and indirect discrimination are connected to effects specific to certain paths. We concentrate on indirect discrimination, even though fulfilling the criterion for direct discrimination is comparable. As mentioned in Section \ref{causal_notation} , $|SE_\pi (s^+, s^-)| = |P (Y_{s^+|\pi_C, s^-|\overline{\pi}_C} ) - P (Y_{s^-} )|<\tau$ must hold for a path set $\pi_C$ that includes paths passing through certain attributes, shown in Fig. \ref{fig:indirect_a} ( in Appendix). $F^1$ for indirect discrimination is similar to that in Section \ref{sec:total}. However, the design of $F^2$ is altered because it needs to adapt to the situation where the intervention is transferred only through $\pi_C$, shown in Fig. \ref{fig:indirect_b} (in Appendix). %To achieve this, we begin by aligning the layout of $F^2$ with the interventional graph $\mathcal{G}s$. %= (V, A \backslash {V_j \rightarrow S}V_j \in P{aS})$.we first make the structure of $F^2$ to match the interventional graph $\mathcal{G}_s$. %= (V, A \backslash \{V_j \rightarrow S\}V_j \in P_{aS})$.
We examine two possible states for the sub-neural network $F^2_S$: the reference state and the interventional state. Under the reference state, $F^2_S$ is constantly set to 0. On the other hand, under the interventional state, $F^2_S$ is set to 1 if $s=s^+$, and 0 if $s=s^-$. For other sub-neural networks, there are also two possible values: the reference state and the interventional state, according to the state of $F^2_S$. If a sub-neural network corresponds to a node that is not present on any path in $\pi_C$, it only accepts reference states as input and generates reference states as output. However, for any other sub-neural network $F^2_{V_j}$ that exists on at least one path in $\pi_C$, it may accept both reference and interventional states as input and generate both types of states as output.%If a sub-neural network corresponds to a node that is not on any path in $\pi_C$, it always takes reference values as input and outputs reference values. However, for any other sub-neural network $F^2_{V_j}$ that is on at least one path in $\pi_C$, it may take both types of values as input and output both. 

\subsubsection{Reweighting for Counterfactual Fairness}
In the context of counterfactual fairness, interventions are made based on a subset of variables $O=o$. %Unlike previous fairness criteria that only consider the interventional model, counterfactual fairness takes into account the relationship between the original causal model and the interventional model. 
%Our proposed method addresses this by establishing a direct dependency between samples in $F^1$ and samples in $F^2$.
 Both $F^1$ and $F^2$ have similar structures to those in Section \ref{sec:indirect}. However, we only use samples in $F^2$ as interventional samples if they satisfy the condition $O=o$. This means that the interventional distribution from $F^2$ is conditioned on $O=o$ as $P_{F^2}(X_s, Y_s|o)$. The discriminator $D$ is designed to distinguish between $\hat{y}{s^+}|o\sim P_{F^2} (Y_{s^+}|o)$ and $\hat{y}{s^-}|o\sim P_{F^2} (Y_{s^-}|o)$, and aims to reach $P_{F^2} (Y_{s^+}|o) = P_{F^2} (Y_{s^-}|o)$. During training, the value of $m$ should be adjusted based on the number of samples that are involved in the intervention. %Note here we only deal with the situation where a specific $o$ represents a subgroup of individuals which we could use them as evidence to update the corresponding neural networks of $F_2$, which is the action step in counterfactual inference. To deal with a individual situation, more assumptions should be introduced here.

 \subsection{Training Algorithm}
To train the network $F^1$ to minimize the loss in Equation \ref{eq:net}, we alternately optimize the network parameters of $F^1$ and $D$ and learn the weights $\mathbf{w}$ by fixing others as known. %We initialize $\mathbf{w}$ by $w_i=1$ for all $i$. Then, we alternately run the following two procedures when training the networks.

\paragraph{Updating parameters of $F^1$ with fixed $\mathbf{w}$} 
Fixing $\mathbf{w}$, we update $F^1$ to minimize the loss in Equation \ref{eq:net} for $M$ steps, using the mini-batch stochastic gradient descent algorithm.

\paragraph{Updating $\mathbf{w}$ with fixed $F^2$ (synchronized with $F^1$)}
Fixing parameters of $F^2$, we control the training data into two groups ($S^+$ and $S^-$) for intervention, and learn $\mathbf{w}$ in Equation~\ref{eq:dis}. Since Equation \ref{eq:dis} is a min-max optimization problem, we can alternately optimize the weights $\mathbf{w}$ and the parameters of $D$ of the discriminator by fixing the other one as known. 
%For reducing the computational cost, we only perform the alternate optimization once, which yields satisfactory performance in experiments. 
Therefore, we first fix $w_i=1$ for all $i$ and optimize $D$ to maximize the objective function in Equation \ref{eq:dis} using the gradient penalty technique, as in WGAN with Gradient Penalty~\cite{gulrajani2017}. Note that when $w_i=1$ for all $i$, Equation \ref{eq:net} is equivalent to the situation when there is no reweighting applied. Then, fixing the discriminator $D$, we optimize $\mathbf{w}$. %We denote $d_i=D(y_i)$ and $\mathbf{d^{s^+}}=(d_1, d_2, ...d_{k})^T$. The optimization problem for $\mathbf{w}$ becomes

%\begin{equation} \label{eq:w}
%\underset{\mathbf{w}}{\text{min}}\,\mathbf{d}^T\mathbf{w},  s.t.  w_i\geqslant0, \sum_{i=1}^{n_p}w_i=n_u,\sum_{i=1}^{n_p}(w_i-1)^2\leqslant Tn_u.
%\end{equation}

We denote $d_k=D(\hat{y}^{S^+}_k)-D(\hat{y}^{S^-}_k)$ and $\mathbf{d}=(d_1, d_2, ...d_m)^T$. The optimization problem for $\mathbf{w}$ becomes a constrained least squares problem:

\begin{equation} \label{eq:w}
\underset{\mathbf{w}}{\text{min}}\,\mathbf{d}^T\mathbf{w},  s.t.  w_k\geqslant0, \sum_{k=1}^{m}w_k=m,\sum_{k=1}^{m}(w_k-1)^2\leqslant Tm
\end{equation}

\begin{comment}
In counterfactual fairness, the intervention is performed conditioning on a subset of variables $O=o$. Thus, different from previous fairness criteria that concern the interventional model only, counterfactual fairness concerns the connection between the original causal model and the interventional model. We reflect this connection in our method by building a direct dependency between the samples through
$F^1$ and the samples through $F^2$. Specifically, the structures of $F^1$ and $F^2$ are similar to those in Section \ref{sec:indirect}. However, for each each sample $(x,y,s)$, and observe whether in the sample we have $O=o$. Only for those samples with $O = o$, we use them in $F^2$ as interventional samples. Thus, the interventional distribution generated by $F^2$ is conditioned on $O=o$, denoted by $P_{F^2}(X_s, Y_s|o)$. Finally, the discriminator $D$ is designed to distinguish between $\hat{y}_{s^+}|o\sim P_{F^2} (Y_{s^+}|o)$
and $\hat{y}_{s^-}|o\sim P_{F^2} (Y_{s^-}|o)$, producing the value function that aims to achieve $P_{F^2} (Y_{s^+}|o) = P_{F^2} (Y_{s^-}|o)$. Then, during training, $m$ should be changed to the corresponding number of samples which are invloved in the intervention. 
\end{comment}

\section{Experimental Evaluation}\label{experiment}
We conduct experiments on two benchmarks datasets (ADULT \cite{kohavi1996} and COMPAS \cite{mattu}) to evaluate our reweighting approach and compare it with state-of-the-art methods: FairGAN \cite{xu2018}, CFGAN \cite{xu2019b} and Causal Inference for Social Discrimination Reasoning (CISD) \cite{qureshi2020} for total effect and indirect discrimination (please refer to Appendix (\ref{sec:dataset}) for more details about the datasets). CISD \cite{qureshi2020} introduces a technique for identifying causal discrimination through the use of propensity score analysis. It consists of mitigating the influence of confounding variables by reweighing samples based on the propensity scores calculated from a logistic regression. The approach, however, is purely statistical with no causal knowledge exploited. We also compare our method with CFGAN and two methods from \cite{kusner2018} (we refer them as $CE_1$ and $CE_3$ in our paper) for counterfactual effect. $CE_1$ only uses on non-descendants of $S$ for classification. $CE_3$ is similar to $CE_1$ but presupposes an additive $U$.
The reason we choose these methods is: FairGAN for statistical parity and CFGAN for causal fairness also use adversarial method to mitigate bias, similar to our design; CISD approaches causal fairness with weighting scheme. We then compare the performance of our method with the mentioned methods on total effect, indirect discrimination and counterfactual fairness with 4 different downstream classifiers: decision tree (DT) \cite{wu2008top}, logistic regression (LR) \cite{cox1958regression}, support vector machine (SVM) \cite{cortes1995support} and random forest (RF) \cite{ho1995random}. We compare the accuracy of the downstream tasks to see if the data preserves good utility, where higher accuracy indicates better utility. For the utility of the downstream task, we also compute the Wasserstein distance between the manipulated data and the original data, where a smaller Wasserstein distance indicates closer the two distributions, and better utility for the downstream tasks.

\subsection{The datasets and setup} 
Due to page limit, please refer to Appendix for the details of datasets and training.

\subsection{Analysis}

\subsubsection{Total Effect}

%We calculate the total effect for the original dataset and manipulated datasets from different methods. The results are shown in Table \ref{table:total}. As can be seen, the original data has a total effect of 0.1854. FairGAN produces no total effect when we apply FairGAN to address demographic parity. In fact, in the design FairGAN, there is no mechanism to deal with causal fairness but more simple statistical fairness notions like demographic parity. However, it does not do well with Wasserstein distance and also in the downstream task. This may be because FairGAN removes too much information. The generated data by CFGAN based on total effect produces no total effect, but it does not perform as well as our method on Wasserstein distance which might be because generated data are not so in distributiona as the reweighted data. We outperform CISD might due to the propensity score used in CISD for reweighting is calculated from logistic regression while our method calculated the weight from the output of a neural network, where we could capture the dataset with higher flexibility.

In Table \ref{table:total}, we present the total effect (TE) calculated for the original dataset and the datasets processed using various methods. The original ADULT dataset has a total effect of 0.1854 and COMPAS 0.2389, while applying FairGAN to achieve demographic parity yields almost no total effect. As mentioned in Section \ref{causal_notation}, total effect is very similar to demographic parity. However, FairGAN is limited by its focus on statistical fairness, rather than causal fairness, and does not perform well on Wasserstein distance or downstream tasks.  It is quite intuitive that if total fairness is met, total fairness should be achieved too on the condition that the causal graph is sufficient. We test it on our two datasets and the result is acceptable. CFGAN produces no total effect, but it performs worse than our method on Wasserstein distance, possibly because reweighted data could manage to stay closer to the original data distribution. Our method also outperforms CISD, which may be due to the use of a neural network instead of logistic regression to calculate weights, allowing for greater flexibility in capturing the dataset.

\begin{figure}
    \centering
    \includegraphics[width=.3\textwidth]{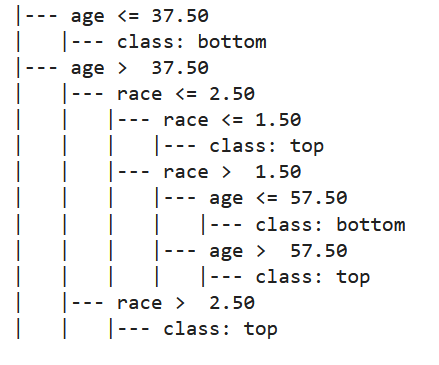}
    \caption{The visualize of the decision tree trying to classify individuals with low or high weights. we see that age and race are the most important attributes to build the tree. The mapping of label encoder for $race$ is $\{'Amer-Indian-Eskimo':0,
 'Asian-Pac-Islander':1,
 'Black':2,
 'Other':3,
 'White':4\}$}
    \label{tree}
\end{figure}

\textit{A Closer Look at the Weights} After ranking the weights of samples in the Adult dataset, we observed that older individuals from Europe or Asia (e.g., Germany and India) tend to have the highest weights, while younger black individuals from Caribbean countries (e.g., Jamaica and Haiti) tend to have lower weights. This suggests that when sex is intervened from female to male, the former group is less influenced by the change, while the latter group is more influenced in terms of income. White, middle-aged individuals born in the US are assigned medium weights. To visualize it, we build a decision tree to classify top 10\% individuals with highest weights and bottom 10\% individuals with lowest weights using the three root nodes $\{race, native\_country, age\}$, shown in Fig. \ref{tree}.  %It is quite intuitive that the employers prefer younger employees and the employees from abroad who are hired in the US are usually with better qualification. We build a decsion tree to visualize the weights and the attributes

\begin{table*}
\centering
\caption{The total effect (TE) and indirect discrimination (SE) on Adult and COMPAS datasets}
\resizebox{\linewidth}{!}{%
\begin{tabular}{|cccllllc|cccclllc|}
\hline
\multicolumn{8}{|c|}{ADULT}                                                                                                                                                                                                                                                                                                                                                                                                                                                                                                                                                                                                                             & \multicolumn{8}{c|}{COMPAS}                                                                                                                                                                                                                                                                                                                                                                                                                                                                                                                                                                                                                             \\ \hline
\multicolumn{1}{|c|}{\multirow{2}{*}{}} & \multicolumn{1}{c|}{\multirow{2}{*}{total effect}}                                                       & \multicolumn{1}{c|}{\multirow{2}{*}{indirect discrimination}}                                  & \multicolumn{1}{c|}{\multirow{2}{*}{Wasserstein distance}}                                             & \multicolumn{4}{c|}{classifier accuracy (\%)}                                                                                                                                                                                                                                                    & \multicolumn{1}{c|}{\multirow{2}{*}{}} & \multicolumn{1}{c|}{\multirow{2}{*}{total effect}}                                                       & \multicolumn{1}{c|}{\multirow{2}{*}{indirect discrimination}}                                            & \multicolumn{1}{c|}{\multirow{2}{*}{Wasserstein distance}}                                              & \multicolumn{4}{c|}{classifier accuracy (\%)}                                                                                                                                                                                                                                                    \\ \cline{5-8} \cline{13-16} 
\multicolumn{1}{|c|}{}                  & \multicolumn{1}{c|}{}                                                                                    & \multicolumn{1}{c|}{}                                                                                    & \multicolumn{1}{c|}{}                                                                & \multicolumn{1}{c|}{SVM}                                                    & \multicolumn{1}{c|}{DT}                                                     & \multicolumn{1}{c|}{LR}                                                     & \multicolumn{1}{c|}{RF}                                & \multicolumn{1}{c|}{}                  & \multicolumn{1}{c|}{}                                                                                    & \multicolumn{1}{c|}{}                                                                                    & \multicolumn{1}{c|}{}                                                                 & \multicolumn{1}{c|}{SVM}                                                    & \multicolumn{1}{c|}{DT}                                                     & \multicolumn{1}{c|}{LR}                                                     & \multicolumn{1}{c|}{RF}                                \\ \hline
\multicolumn{1}{|c|}{original data}         & \multicolumn{1}{c|}{\begin{tabular}[c]{@{}c@{}}0.1854\\  (0.0301)\end{tabular}}                          & \multicolumn{1}{c|}{\begin{tabular}[c]{@{}c@{}}0.1773\\ (0.0489)\end{tabular}}                           & \multicolumn{1}{c|}{0}                                                               & \multicolumn{1}{c|}{\begin{tabular}[c]{@{}l@{}}81.78\\ (1.45)\end{tabular}} & \multicolumn{1}{c|}{\begin{tabular}[c]{@{}l@{}}81.77\\ (1.75)\end{tabular}} & \multicolumn{1}{c|}{\begin{tabular}[c]{@{}l@{}}81.70\\ (1.63)\end{tabular}} & \begin{tabular}[c]{@{}l@{}}81.78\\ (1.76)\end{tabular} & \multicolumn{1}{c|}{orginal data}         & \multicolumn{1}{c|}{\begin{tabular}[c]{@{}c@{}}0.2389\\ (0.0245)\end{tabular}}                           & \multicolumn{1}{c|}{\begin{tabular}[c]{@{}c@{}}0.2137\\ (0.0985)\end{tabular}}                           & \multicolumn{1}{c|}{0}                                                                & \multicolumn{1}{c|}{\begin{tabular}[c]{@{}l@{}}65.24\\ (2.34)\end{tabular}} & \multicolumn{1}{c|}{\begin{tabular}[c]{@{}l@{}}65.15\\ (1.46)\end{tabular}} & \multicolumn{1}{c|}{\begin{tabular}[c]{@{}l@{}}65.10\\ (2.19)\end{tabular}} & \begin{tabular}[c]{@{}l@{}}65.27\\ (1.09)\end{tabular} \\ \hline
\multicolumn{1}{|c|}{Ours (TE)}          & \multicolumn{1}{c|}{\multirow{2}{*}{\textbf{\begin{tabular}[c]{@{}c@{}}0.0017\\ (0.0009)\end{tabular}}}} & \multicolumn{1}{c|}{\multirow{2}{*}{\textbf{\begin{tabular}[c]{@{}c@{}}0.0012\\ (0.0007)\end{tabular}}}} & \multicolumn{1}{c|}{\textbf{\begin{tabular}[c]{@{}c@{}}0.71\\ (0.19)\end{tabular}}} & \multicolumn{1}{c|}{\begin{tabular}[c]{@{}c@{}}81.12\\ (1.72)\end{tabular}} & \multicolumn{1}{c|}{\begin{tabular}[c]{@{}l@{}}81.20\\ (1.86)\end{tabular}} & \multicolumn{1}{c|}{\begin{tabular}[c]{@{}c@{}}81.60\\ (2.03)\end{tabular}} & \begin{tabular}[c]{@{}l@{}}81.14\\ (1.05)\end{tabular} & \multicolumn{1}{c|}{Ours (TE)}          & \multicolumn{1}{c|}{\multirow{2}{*}{\textbf{\begin{tabular}[c]{@{}c@{}}0.0037\\ (0.0018)\end{tabular}}}} & \multicolumn{1}{c|}{\multirow{2}{*}{\begin{tabular}[c]{@{}c@{}}0.0017\\ (0.0009)\end{tabular}}}          & \multicolumn{1}{c|}{\textbf{\begin{tabular}[c]{@{}c@{}}1.21\\ (0.32)\end{tabular}}}  & \multicolumn{1}{c|}{\begin{tabular}[c]{@{}l@{}}65.09\\ (2.75)\end{tabular}} & \multicolumn{1}{c|}{\begin{tabular}[c]{@{}l@{}}65.13\\ (1.76)\end{tabular}} & \multicolumn{1}{c|}{\begin{tabular}[c]{@{}l@{}}65.06\\ (2.08)\end{tabular}} & \begin{tabular}[c]{@{}l@{}}65.11\\ (1.02)\end{tabular} \\ \cline{1-1} \cline{4-9} \cline{12-16} 
\multicolumn{1}{|c|}{Ours (SE)}          & \multicolumn{1}{c|}{}                                                                                    & \multicolumn{1}{c|}{}                                                                                    & \multicolumn{1}{c|}{\textbf{\begin{tabular}[c]{@{}c@{}}0.69\\ (0.23)\end{tabular}}} & \multicolumn{1}{c|}{\begin{tabular}[c]{@{}l@{}}81.14\\ (1.58)\end{tabular}} & \multicolumn{1}{c|}{\begin{tabular}[c]{@{}l@{}}80.97\\ (2.01)\end{tabular}} & \multicolumn{1}{c|}{\begin{tabular}[c]{@{}l@{}}81.65\\ (1.96)\end{tabular}} & \begin{tabular}[c]{@{}l@{}}81.17\\ (1.92)\end{tabular} & \multicolumn{1}{c|}{Ours (SE)}          & \multicolumn{1}{c|}{}                                                                                    & \multicolumn{1}{c|}{}                                                                                    & \multicolumn{1}{c|}{\textbf{\begin{tabular}[c]{@{}c@{}}0.72\\ (0.35)\end{tabular}}} & \multicolumn{1}{c|}{\begin{tabular}[c]{@{}c@{}}65.11\\ (1.98)\end{tabular}} & \multicolumn{1}{c|}{\begin{tabular}[c]{@{}l@{}}65.14\\ (2.06)\end{tabular}} & \multicolumn{1}{c|}{\begin{tabular}[c]{@{}l@{}}65.02\\ (1.12)\end{tabular}} & \begin{tabular}[c]{@{}l@{}}65.09\\ (1.95)\end{tabular} \\ \hline
\multicolumn{1}{|c|}{FairGAN}           & \multicolumn{1}{c|}{\begin{tabular}[c]{@{}c@{}}0.0021\\ (0.0007)\end{tabular}}                           & \multicolumn{1}{c|}{\begin{tabular}[c]{@{}c@{}}0.0148\\ (0.0075)\end{tabular}}                           & \multicolumn{1}{c|}{\begin{tabular}[c]{@{}c@{}}5.21\\ (0.78)\end{tabular}}          & \multicolumn{1}{c|}{\begin{tabular}[c]{@{}l@{}}79.88\\ (1.47)\end{tabular}} & \multicolumn{1}{c|}{\begin{tabular}[c]{@{}l@{}}79.81\\ (1.89)\end{tabular}} & \multicolumn{1}{c|}{\begin{tabular}[c]{@{}l@{}}80.36\\ (1.32)\end{tabular}} & \begin{tabular}[c]{@{}l@{}}80.82\\ (1.65)\end{tabular} & \multicolumn{1}{c|}{FairGAN}           & \multicolumn{1}{c|}{\begin{tabular}[c]{@{}c@{}}0.0075\\ (0.0056)\end{tabular}}                           & \multicolumn{1}{c|}{\begin{tabular}[c]{@{}c@{}}0.0341\\ (0.0075)\end{tabular}}                           & \multicolumn{1}{c|}{\begin{tabular}[c]{@{}c@{}}3.24\\ (1.45)\end{tabular}}          & \multicolumn{1}{c|}{\begin{tabular}[c]{@{}l@{}}64.24\\ (1.77)\end{tabular}} & \multicolumn{1}{c|}{\begin{tabular}[c]{@{}l@{}}64.15\\ (2.01)\end{tabular}} & \multicolumn{1}{c|}{\begin{tabular}[c]{@{}l@{}}64.50\\ (2.75)\end{tabular}} & \begin{tabular}[c]{@{}l@{}}64.26\\ (2.34)\end{tabular} \\ \hline
\multicolumn{1}{|c|}{CFGAN (TE)}        & \multicolumn{1}{c|}{\multirow{2}{*}{\begin{tabular}[c]{@{}c@{}}0.0106\\ (0.0008)\end{tabular}}}          & \multicolumn{1}{c|}{\multirow{2}{*}{\begin{tabular}[c]{@{}c@{}}0.0034\\ (0.0012)\end{tabular}}}          & \multicolumn{1}{c|}{\begin{tabular}[c]{@{}c@{}}1.78\\ (0.65)\end{tabular}}          & \multicolumn{1}{c|}{\begin{tabular}[c]{@{}l@{}}80.34\\ (2.56)\end{tabular}} & \multicolumn{1}{c|}{\begin{tabular}[c]{@{}l@{}}80.15\\ (1.52)\end{tabular}} & \multicolumn{1}{c|}{\begin{tabular}[c]{@{}l@{}}80.07\\ (1.65)\end{tabular}} & \begin{tabular}[c]{@{}l@{}}80.39\\ (1.32)\end{tabular} & \multicolumn{1}{c|}{CFGAN (TE)}        & \multicolumn{1}{c|}{\multirow{2}{*}{\begin{tabular}[c]{@{}c@{}}0.0364\\ (0.0175)\end{tabular}}}          & \multicolumn{1}{c|}{\multirow{2}{*}{\textbf{\begin{tabular}[c]{@{}c@{}}0.0016\\ (0.0025)\end{tabular}}}} & \multicolumn{1}{c|}{\begin{tabular}[c]{@{}c@{}}2.76\\ (1.65)\end{tabular}}          & \multicolumn{1}{c|}{\begin{tabular}[c]{@{}l@{}}64.59\\ (2.65)\end{tabular}} & \multicolumn{1}{c|}{\begin{tabular}[c]{@{}l@{}}65.13\\ (2.73)\end{tabular}} & \multicolumn{1}{c|}{\begin{tabular}[c]{@{}l@{}}65.02\\ (2.03)\end{tabular}} & \begin{tabular}[c]{@{}l@{}}65.01\\ (2.45)\end{tabular} \\ \cline{1-1} \cline{4-9} \cline{12-16} 
\multicolumn{1}{|c|}{CFGAN (SE)}        & \multicolumn{1}{c|}{}                                                                                    & \multicolumn{1}{c|}{}                                                                                    & \multicolumn{1}{c|}{\begin{tabular}[c]{@{}c@{}}1.89\\ (0.29)\end{tabular}}          & \multicolumn{1}{c|}{\begin{tabular}[c]{@{}l@{}}80.37\\ (1.56)\end{tabular}} & \multicolumn{1}{c|}{\begin{tabular}[c]{@{}l@{}}80.49\\ (2.05)\end{tabular}} & \multicolumn{1}{c|}{\begin{tabular}[c]{@{}l@{}}80.04\\ (1.67)\end{tabular}} & \begin{tabular}[c]{@{}l@{}}80.24\\ (1.09)\end{tabular} & \multicolumn{1}{c|}{CFGAN (SE)}        & \multicolumn{1}{c|}{}                                                                                    & \multicolumn{1}{c|}{}                                                                                    & \multicolumn{1}{c|}{\begin{tabular}[c]{@{}c@{}}2.64\\ (0.91)\end{tabular}}           & \multicolumn{1}{c|}{\begin{tabular}[c]{@{}l@{}}64.21\\ (2.45)\end{tabular}} & \multicolumn{1}{c|}{\begin{tabular}[c]{@{}l@{}}64.25\\ (1.75)\end{tabular}} & \multicolumn{1}{c|}{\begin{tabular}[c]{@{}l@{}}64.80\\ (1.97)\end{tabular}} & \begin{tabular}[c]{@{}l@{}}64.87\\ (1.54)\end{tabular} \\ \hline
\multicolumn{1}{|c|}{CISD (TE)}          & \multicolumn{1}{c|}{\multirow{2}{*}{\begin{tabular}[c]{@{}c@{}}0.0206\\ (0.0074)\end{tabular}}}          & \multicolumn{1}{c|}{\multirow{2}{*}{\begin{tabular}[c]{@{}c@{}}0.0098\\ (0.0045)\end{tabular}}}          & \multicolumn{1}{c|}{\begin{tabular}[c]{@{}c@{}}2.57\\ (0.18)\end{tabular}}          & \multicolumn{1}{c|}{\begin{tabular}[c]{@{}l@{}}80.73\\ (1.42)\end{tabular}} & \multicolumn{1}{c|}{\begin{tabular}[c]{@{}l@{}}80.74\\ (1.75)\end{tabular}} & \multicolumn{1}{c|}{\begin{tabular}[c]{@{}l@{}}81.15\\ (1.82)\end{tabular}} & \begin{tabular}[c]{@{}l@{}}81.27\\ (1.47)\end{tabular} & \multicolumn{1}{c|}{CISD (TE)}          & \multicolumn{1}{c|}{\multirow{2}{*}{\begin{tabular}[c]{@{}c@{}}0.0356\\ (0.0246)\end{tabular}}}          & \multicolumn{1}{c|}{\multirow{2}{*}{\begin{tabular}[c]{@{}c@{}}0.0175\\ (0.0231)\end{tabular}}}          & \multicolumn{1}{c|}{\begin{tabular}[c]{@{}c@{}}2.57\\ (1.61)\end{tabular}}          & \multicolumn{1}{c|}{\begin{tabular}[c]{@{}l@{}}65.04\\ (1.76)\end{tabular}} & \multicolumn{1}{c|}{\begin{tabular}[c]{@{}l@{}}65.17\\ (1.54)\end{tabular}} & \multicolumn{1}{c|}{\begin{tabular}[c]{@{}l@{}}65.04\\ (2.47)\end{tabular}} & \begin{tabular}[c]{@{}l@{}}65.05\\ (1.75)\end{tabular} \\ \cline{1-1} \cline{4-9} \cline{12-16} 
\multicolumn{1}{|c|}{CISD (SE)}          & \multicolumn{1}{c|}{}                                                                                    & \multicolumn{1}{c|}{}                                                                                    & \multicolumn{1}{c|}{\begin{tabular}[c]{@{}c@{}}2.82\\ (0.23)\end{tabular}}          & \multicolumn{1}{c|}{\begin{tabular}[c]{@{}l@{}}80.75\\ (1.28)\end{tabular}} & \multicolumn{1}{c|}{\begin{tabular}[c]{@{}l@{}}80.72\\ (1.58)\end{tabular}} & \multicolumn{1}{c|}{\begin{tabular}[c]{@{}l@{}}80.77\\ (1.96)\end{tabular}} & \begin{tabular}[c]{@{}l@{}}81.32\\ (1.95)\end{tabular} & \multicolumn{1}{c|}{CISD (SE)}          & \multicolumn{1}{c|}{}                                                                                    & \multicolumn{1}{c|}{}                                                                                    & \multicolumn{1}{c|}{\begin{tabular}[c]{@{}c@{}}2.65\\ (1.56)\end{tabular}}          & \multicolumn{1}{c|}{\begin{tabular}[c]{@{}l@{}}64.01\\ (1.56)\end{tabular}} & \multicolumn{1}{c|}{\begin{tabular}[c]{@{}l@{}}65.02\\ (1.49)\end{tabular}} & \multicolumn{1}{c|}{\begin{tabular}[c]{@{}l@{}}64.09\\ (2.45)\end{tabular}} & \begin{tabular}[c]{@{}l@{}}64.11\\ (1.32)\end{tabular} \\ \hline
\end{tabular}}\label{table:total}
\end{table*}

\begin{table*}
\centering
\caption{The counterfactual effect (CE) on Adult and COMPAS datasets}
\resizebox{\linewidth}{!}{%
\begin{tabular}{|cccccccc|cccccccc|}
\hline
\multicolumn{8}{|c|}{ADULT}                                                                                                                                                                                                                                                                                                                                                                                                                                                                                                                                                                                          & \multicolumn{8}{c|}{COMPAS}                                                                                                                                                                                                                                                                                                                                                                                                                                                                                                                                                                                          \\ \hline
\multicolumn{1}{|c|}{\multirow{2}{*}{}} & \multicolumn{2}{c|}{counterfactual effect}                                                                                                                                        & \multicolumn{1}{c|}{\multirow{2}{*}{Wasserstein distance}}                                            & \multicolumn{4}{c|}{classifier accuracy (\%)}                                                                                                                                                                                                                                                    & \multicolumn{1}{c|}{\multirow{2}{*}{}} & \multicolumn{2}{c|}{counterfactual effect}                                                                                                                                        & \multicolumn{1}{c|}{\multirow{2}{*}{Wasserstein distance}}                                             & \multicolumn{4}{c|}{classifier accuracy (\%)}                                                                                                                                                                                                                                                    \\ \cline{2-3} \cline{5-8} \cline{10-11} \cline{13-16} 
\multicolumn{1}{|c|}{}                  & \multicolumn{1}{c|}{$o_1$}                                                                 & \multicolumn{1}{c|}{$o_2$}                                                                 & \multicolumn{1}{c|}{}                                                               & \multicolumn{1}{c|}{SVM}                                                    & \multicolumn{1}{c|}{DT}                                                     & \multicolumn{1}{c|}{LR}                                                     & RF                                                     & \multicolumn{1}{c|}{}                  & \multicolumn{1}{c|}{$o_1$}                                                                 & \multicolumn{1}{c|}{$o_2$}                                                                 & \multicolumn{1}{c|}{}                                                                & \multicolumn{1}{c|}{SVM}                                                    & \multicolumn{1}{c|}{DT}                                                     & \multicolumn{1}{c|}{LR}                                                     & RF                                                     \\ \hline
\multicolumn{1}{|c|}{original data}     & \multicolumn{1}{c|}{\begin{tabular}[c]{@{}c@{}}0.1254\\  (0.0107)\end{tabular}}         & \multicolumn{1}{c|}{\begin{tabular}[c]{@{}c@{}}0.1265\\ (0.0489)\end{tabular}}          & \multicolumn{1}{c|}{0}                                                              & \multicolumn{1}{c|}{81.78}                                                  & \multicolumn{1}{c|}{81.77}                                                  & \multicolumn{1}{c|}{81.70}                                                  & 81.78                                                  & \multicolumn{1}{c|}{original data}     & \multicolumn{1}{c|}{\begin{tabular}[c]{@{}c@{}}0.2079\\  (0.0303)\end{tabular}}         & \multicolumn{1}{c|}{\begin{tabular}[c]{@{}c@{}}0.1973\\ (0.0484)\end{tabular}}          & \multicolumn{1}{c|}{0}                                                               & \multicolumn{1}{c|}{\begin{tabular}[c]{@{}c@{}}65.24\\ (2.35)\end{tabular}} & \multicolumn{1}{c|}{\begin{tabular}[c]{@{}c@{}}65.15\\ (1.32)\end{tabular}} & \multicolumn{1}{c|}{\begin{tabular}[c]{@{}c@{}}65.10\\ (1.97)\end{tabular}} & \begin{tabular}[c]{@{}c@{}}65.27\\ (1.35)\end{tabular} \\ \hline
\multicolumn{1}{|c|}{Ours}              & \multicolumn{1}{c|}{\textbf{\begin{tabular}[c]{@{}c@{}}0.0017\\ (0.0009)\end{tabular}}} & \multicolumn{1}{c|}{\begin{tabular}[c]{@{}c@{}}0.0030\\ (0.0007)\end{tabular}}          & \multicolumn{1}{c|}{\textbf{\begin{tabular}[c]{@{}c@{}}0.98\\ (0.10)\end{tabular}}} & \multicolumn{1}{c|}{\begin{tabular}[c]{@{}c@{}}81.14\\ (1.32)\end{tabular}} & \multicolumn{1}{c|}{\begin{tabular}[c]{@{}c@{}}81.25\\ (1.37)\end{tabular}} & \multicolumn{1}{c|}{\begin{tabular}[c]{@{}c@{}}81.67\\ (1.36)\end{tabular}} & \begin{tabular}[c]{@{}c@{}}81.29\\ (1.42)\end{tabular} & \multicolumn{1}{c|}{Ours}              & \multicolumn{1}{c|}{\textbf{\begin{tabular}[c]{@{}c@{}}0.0027\\ (0.0009)\end{tabular}}} & \multicolumn{1}{c|}{\begin{tabular}[c]{@{}c@{}}0.0037\\ (0.0109)\end{tabular}}          & \multicolumn{1}{c|}{\textbf{\begin{tabular}[c]{@{}c@{}}1.43\\ (1.68)\end{tabular}}} & \multicolumn{1}{c|}{\begin{tabular}[c]{@{}c@{}}65.11\\ (2.01)\end{tabular}} & \multicolumn{1}{c|}{\begin{tabular}[c]{@{}c@{}}65.14\\ (1.35)\end{tabular}} & \multicolumn{1}{c|}{\begin{tabular}[c]{@{}c@{}}65.07\\ (1.89)\end{tabular}} & \begin{tabular}[c]{@{}c@{}}65.13\\ (1.68)\end{tabular} \\ \hline
\multicolumn{1}{|c|}{$CE_1$}               & \multicolumn{1}{c|}{\begin{tabular}[c]{@{}c@{}}0.0021\\ (0.0157)\end{tabular}}          & \multicolumn{1}{c|}{\begin{tabular}[c]{@{}c@{}}0.0148\\ (0.0025)\end{tabular}}          & \multicolumn{1}{c|}{\begin{tabular}[c]{@{}c@{}}4.95\\ (0.45)\end{tabular}}         & \multicolumn{1}{c|}{\begin{tabular}[c]{@{}c@{}}77.88\\ (1.65)\end{tabular}} & \multicolumn{1}{c|}{\begin{tabular}[c]{@{}c@{}}78.81\\ (1.32)\end{tabular}} & \multicolumn{1}{c|}{\begin{tabular}[c]{@{}c@{}}76.36\\ (2.06)\end{tabular}} & \begin{tabular}[c]{@{}c@{}}77.82\\ (1.78)\end{tabular} & \multicolumn{1}{c|}{$CE_1$}               & \multicolumn{1}{c|}{\begin{tabular}[c]{@{}c@{}}0.0034\\ (0.0007)\end{tabular}}          & \multicolumn{1}{c|}{\begin{tabular}[c]{@{}c@{}}0.0048\\ (0.0075)\end{tabular}}          & \multicolumn{1}{c|}{\begin{tabular}[c]{@{}c@{}}4.53\\ (1.56)\end{tabular}}         & \multicolumn{1}{c|}{\begin{tabular}[c]{@{}c@{}}63.24\\ (1.76)\end{tabular}} & \multicolumn{1}{c|}{\begin{tabular}[c]{@{}c@{}}63.16\\ (1.43)\end{tabular}} & \multicolumn{1}{c|}{\begin{tabular}[c]{@{}c@{}}62.19\\ (1.46)\end{tabular}} & \begin{tabular}[c]{@{}c@{}}62.27\\ (2.21)\end{tabular} \\ \hline
\multicolumn{1}{|c|}{$CE_3$}               & \multicolumn{1}{c|}{\begin{tabular}[c]{@{}c@{}}0.1123\\ (0.0017)\end{tabular}}          & \multicolumn{1}{c|}{\begin{tabular}[c]{@{}c@{}}0.1046\\ (0.0057)\end{tabular}}          & \multicolumn{1}{c|}{\begin{tabular}[c]{@{}c@{}}3.11\\ (0.65)\end{tabular}}         & \multicolumn{1}{c|}{\begin{tabular}[c]{@{}c@{}}80.75\\ (1.76)\end{tabular}} & \multicolumn{1}{c|}{\begin{tabular}[c]{@{}c@{}}81.59\\ (2.15)\end{tabular}} & \multicolumn{1}{c|}{\begin{tabular}[c]{@{}c@{}}81.62\\ (2.01)\end{tabular}} & \begin{tabular}[c]{@{}c@{}}81.43\\ (1.37)\end{tabular} & \multicolumn{1}{c|}{$CE_3$}               & \multicolumn{1}{c|}{\begin{tabular}[c]{@{}c@{}}0.1021\\ (0.0981)\end{tabular}}          & \multicolumn{1}{c|}{\begin{tabular}[c]{@{}c@{}}0.0145\\ (0.0025)\end{tabular}}          & \multicolumn{1}{c|}{\begin{tabular}[c]{@{}c@{}}1.76\\ (0.89)\end{tabular}}          & \multicolumn{1}{c|}{\begin{tabular}[c]{@{}c@{}}64.07\\ (1.96)\end{tabular}} & \multicolumn{1}{c|}{\begin{tabular}[c]{@{}c@{}}65.03\\ (1.95)\end{tabular}} & \multicolumn{1}{c|}{\begin{tabular}[c]{@{}c@{}}64.01\\ (1.45)\end{tabular}} & \begin{tabular}[c]{@{}c@{}}65.10\\ (2.89)\end{tabular} \\ \hline
\multicolumn{1}{|c|}{CFGAN (CE)}         & \multicolumn{1}{c|}{\begin{tabular}[c]{@{}c@{}}0.0021\\ (0.0039)\end{tabular}}          & \multicolumn{1}{c|}{\textbf{\begin{tabular}[c]{@{}c@{}}0.0027\\ (0.0064)\end{tabular}}} & \multicolumn{1}{c|}{\begin{tabular}[c]{@{}c@{}}2.99\\ (0.78)\end{tabular}}         & \multicolumn{1}{c|}{\begin{tabular}[c]{@{}c@{}}81.03\\ (1.35)\end{tabular}} & \multicolumn{1}{c|}{\begin{tabular}[c]{@{}c@{}}81.14\\ (2.19)\end{tabular}} & \multicolumn{1}{c|}{\begin{tabular}[c]{@{}c@{}}81.11\\ (2.17)\end{tabular}} & \begin{tabular}[c]{@{}c@{}}81.15\\ (1.95)\end{tabular} & \multicolumn{1}{c|}{CFGAN(CE)}         & \multicolumn{1}{c|}{\begin{tabular}[c]{@{}c@{}}0.0034\\ (0.0024)\end{tabular}}          & \multicolumn{1}{c|}{\textbf{\begin{tabular}[c]{@{}c@{}}0.0031\\ (0.0045)\end{tabular}}} & \multicolumn{1}{c|}{\begin{tabular}[c]{@{}c@{}}1.74\\ (1.06)\end{tabular}}         & \multicolumn{1}{c|}{\begin{tabular}[c]{@{}c@{}}65.04\\ (1.72)\end{tabular}} & \multicolumn{1}{c|}{\begin{tabular}[c]{@{}c@{}}65.11\\ (1.74)\end{tabular}} & \multicolumn{1}{c|}{\begin{tabular}[c]{@{}c@{}}64.08\\ (1.19)\end{tabular}} & \begin{tabular}[c]{@{}c@{}}64.29\\ (1.32)\end{tabular} \\ \hline
\end{tabular}}\label{table:ce}
\end{table*}

\subsubsection{Indirect Discrimination}

%For indirect discrimination, we consider all the paths passing
%through marital status as the path $\pi_C$. The results are also shown in
%Table \ref{table:total}. Similar to total effect, FairGAN removes indirect discrimination but causes the largest utility loss.

To address indirect discrimination (SE), we identify all possible paths except the direct one $\{S\shortrightarrow Y \}$ as the path $\pi_C$ and evaluated the results in Table \ref{table:total}. Similar to total effect, FairGAN removes indirect discrimination but at the cost of significant utility loss. %\textcolor{red}{not so sure how to justify it because FAIRGAN only deals with demographic parity, does total effect has anything to do with demographic parity, or can we claim that if demographic parity it met, total effect is met or the other way around?} 
In contrast, both CISD and our method can effectively remove indirect discrimination while maintaining better data utility than FairGAN. Although CFGAN and CISD perform similarly using different techniques, our method outperforms both methods in terms of Wasserstein distance, indicating the best overall utility among these approaches.
%On the other hand, CISD and our method can remove indirect discrimination and also have better data utility than FAIRGAN. We see that FAIRGAN and CISD achieve comparable performance based on different techniques. However, our method performs better on the dimension of Wasserstein distance which indicates the best utility among these methods. 

\subsubsection{Counterfactual Fairness}
%For counterfactual fairness, we consider the observation of
%two attributes, $O = \{race, %native\_country\}$ for ADULT (we binarize these two attributes), and $O = \{sex, age\}$ for COMPAS. We have 4 value combinations. Table \ref{table:ce} shows the results for 2 subgroups due to space restriction (please refer to Appendix to more details). We could see that the original data contains biases in terms of counterfactual fairness in all subgroups. $A_1$ is counterfactual fair as expected since it is proved to be so in \cite{kusner2018}. However, the data utility is bad especially in terms of classifier accuracy, since it only uses non-descendants of the sensitive attributes for outcome attributes. $A_3$ cannot achieve counterfactual fairness, probably because its linear assumption does not fit the original data well. Finally, our method achieves both counterfactual fairness and good data utility. CFGAN does not performs so well in Wasserstein distance as our method probably because reweighting is doing better at staying close to original distribution than generating methods.

To evaluate counterfactual effect (CE), we consider the conditions on two variables - race and native country (binarized) for ADULT, and sex and age (binarized) for COMPAS - resulting in four value combinations. Table \ref{table:ce} presents the results for two selections (see Appendix (\ref{counterfactual}) for more details). We find biases in the original data regarding counterfactual fairness in these two selections. $CE_1$ is counterfactually fair, but the classifier accuracy is poor because it solely employs non-descendants of the sensitive attributes for outcome attributes. $CE_3$ cannot achieve counterfactual fairness, probably due to the strong assumptions while introducing $U$. In contrast, our method performs well on both dimensions due to its flexibility. Although CFGAN performs well in some aspects, our method outperforms it in Wasserstein distance, likely because reweighting better preserves the original distribution than generation methods. 

\textbf{Summary} We find out that in general neural nets-based methods outperform due to the flexibility of neural networks to capture any function, while reweighting outperforms generation. We could see from the experiment results above, imposing strong assumptions on the $U$ and $F$ could cause unwanted problems, and we argue that is why neural nets should be explored more in causal fairness problem settings. Fairness related methods usually formalize the problem as an optimization trade-off between utility and specific fairness objectives. Nevertheless, these discussions are often based on a fixed distribution that does not align with our current situation. We think that an ideal distribution might exist where fairness and utility are in harmony. To include the reweighting scheme into the downstream tasks could be an very interesting future direction to locate this harmonious distribution.

\section{Conclusion, Limitation and Future Work}\label{conclusion}

We propose a novel approach for achieving causal fairness by dataset reweighting. Our method considers different causal fairness objectives, such as total fairness, path-specific fairness and counterfactual fairness. It consists of two feed-forward neural networks $F^1$ and $F^2$ and a discriminator $D$. The structures of $F^1$ and $F^2$ are designed based on the original causal graph $\mathcal{G}$ and interventional graph $\mathcal{G}_s$, and the discriminator $D$ is used to ensure causal fairness combined with a reweighting scheme. Our experiments on two datasets show that the approach improves over state-of-the-art approaches for the considered causal fairness notions achieving minimal loss of utility. Moreover, by analyzing the sample weights assigned by the approach, the user can gain an understanding of the distribution of the biases in the original dataset. Future work involve analyzing the sample weights further, e.g., by using methods from the eXplainable in AI research area. As another relevant research direction, since practitioners often lack sufficient causal graphs when working with a dataset \cite{binkyte-sadauskiene2022}, an extension of our work could involve causal discovery as an integral part of the approach.

\section{Acknowledgement}
This work has received funding from the European Union’s Horizon 2020 research and innovation programme under Marie Sklodowska-Curie Actions (grant agreement number 860630) for the project “NoBIAS - Artificial Intelligence without Bias”.

\begin{comment}
We proposed the causal fairness-aware reweighting scheme for high quality reweighted fair data.
We considered various causal-based fairness criteria, including total effect, direct discrimination, indirect discrimination, and counterfactual fairness. Our method consists of two (sequential) neural networks to approximate the causal graph and interventional graph with help of a discriminator to guarantee causal fairness. The two neural netowrks aim to simulate the original causal model and the interventional model. This is achieved by arranging the neural network structure following the original causal graph and the interventional graph. Then, the discriminator are adopted for achieving causal fairness. Experiments on different datasets showed that our method can achieve all types of fairness with relatively small utility loss. By having a closer look at the higher and lower samples, we could gain a better understanding of the datasets. However, it might be possible for us to further analize the sample weights as future work. We also have to face the fact that usually when the practitioners deal with a dataset, they usually could not get a sufficient causal graph \cite{binkyte-sadauskiene2022}, how to include causal discovery in our method could be a potential extension of our work. 
\end{comment}

\bibliographystyle{IEEEtran}  
\bibliography{references}

\begin{thebibliography}{10}
\providecommand{\url}[1]{#1}
\csname url@samestyle\endcsname
\providecommand{\newblock}{\relax}
\providecommand{\bibinfo}[2]{#2}
\providecommand{\BIBentrySTDinterwordspacing}{\spaceskip=0pt\relax}
\providecommand{\BIBentryALTinterwordstretchfactor}{4}
\providecommand{\BIBentryALTinterwordspacing}{\spaceskip=\fontdimen2\font plus
\BIBentryALTinterwordstretchfactor\fontdimen3\font minus
  \fontdimen4\font\relax}
\providecommand{\BIBforeignlanguage}[2]{{%
\expandafter\ifx\csname l@#1\endcsname\relax
\typeout{** WARNING: IEEEtran.bst: No hyphenation pattern has been}%
\typeout{** loaded for the language `#1'. Using the pattern for}%
\typeout{** the default language instead.}%
\else
\language=\csname l@#1\endcsname
\fi
#2}}
\providecommand{\BIBdecl}{\relax}
\BIBdecl

\bibitem{pedreshi2008}
D.~Pedreschi, S.~Ruggieri, and F.~Turini, ``Discrimination-aware data mining,''
  in \emph{{KDD}}.\hskip 1em plus 0.5em minus 0.4em\relax {ACM}, 2008, pp.
  560--568.

\bibitem{zliobaite2011}
I.~Zliobaite, F.~Kamiran, and T.~Calders, ``Handling conditional
  discrimination,'' in \emph{{ICDM}}.\hskip 1em plus 0.5em minus 0.4em\relax
  {IEEE} Computer Society, 2011, pp. 992--1001.

\bibitem{hardt2016a}
M.~Hardt, E.~Price, and N.~Srebro, ``Equality of opportunity in supervised
  learning,'' in \emph{{NIPS}}, 2016, pp. 3315--3323.

\bibitem{zhang2016}
L.~Zhang, Y.~Wu, and X.~Wu, ``A causal framework for discovering and removing
  direct and indirect discrimination,'' in \emph{{IJCAI}}.\hskip 1em plus 0.5em
  minus 0.4em\relax ijcai.org, 2017, pp. 3929--3935.

\bibitem{zhang2018}
------, ``Achieving non-discrimination in prediction,'' in
  \emph{{IJCAI}}.\hskip 1em plus 0.5em minus 0.4em\relax ijcai.org, 2018, pp.
  3097--3103.

\bibitem{feldman2015}
M.~Feldman, S.~A. Friedler, J.~Moeller, C.~Scheidegger, and
  S.~Venkatasubramanian, ``Certifying and removing disparate impact,'' in
  \emph{{KDD}}.\hskip 1em plus 0.5em minus 0.4em\relax {ACM}, 2015, pp.
  259--268.

\bibitem{zhang2019a}
L.~Zhang, Y.~Wu, and X.~Wu, ``Causal modeling-based discrimination discovery
  and removal: Criteria, bounds, and algorithms,'' \emph{{IEEE} Trans. Knowl.
  Data Eng.}, vol.~31, no.~11, pp. 2035--2050, 2019.

\bibitem{edwards2016}
H.~Edwards and A.~J. Storkey, ``Censoring representations with an adversary,''
  in \emph{{ICLR} (Poster)}, 2016.

\bibitem{xie2018}
Q.~Xie, Z.~Dai, Y.~Du, E.~H. Hovy, and G.~Neubig, ``Controllable invariance
  through adversarial feature learning,'' in \emph{{NIPS}}, 2017, pp. 585--596.

\bibitem{madras2018c}
D.~Madras, E.~Creager, T.~Pitassi, and R.~S. Zemel, ``Learning adversarially
  fair and transferable representations,'' in \emph{{ICML}}, ser. Proceedings
  of Machine Learning Research, vol.~80.\hskip 1em plus 0.5em minus 0.4em\relax
  {PMLR}, 2018, pp. 3381--3390.

\bibitem{zhang2018a}
B.~H. Zhang, B.~Lemoine, and M.~Mitchell, ``Mitigating unwanted biases with
  adversarial learning,'' in \emph{{AIES}}.\hskip 1em plus 0.5em minus
  0.4em\relax {ACM}, 2018, pp. 335--340.

\bibitem{xu2019b}
D.~Xu, Y.~Wu, S.~Yuan, L.~Zhang, and X.~Wu, ``Achieving causal fairness through
  generative adversarial networks,'' in \emph{{IJCAI}}.\hskip 1em plus 0.5em
  minus 0.4em\relax ijcai.org, 2019, pp. 1452--1458.

\bibitem{roh2021a}
Y.~Roh, K.~Lee, S.~Whang, and C.~Suh, ``Sample selection for fair and robust
  training,'' in \emph{NeurIPS}, 2021, pp. 815--827.

\bibitem{calmon2017a}
F.~P. Calmon, D.~Wei, B.~Vinzamuri, K.~N. Ramamurthy, and K.~R. Varshney,
  ``Optimized pre-processing for discrimination prevention,'' in \emph{{NIPS}},
  2017, pp. 3992--4001.

\bibitem{aghaei2019a}
S.~Aghaei, M.~J. Azizi, and P.~Vayanos, ``Learning optimal and fair decision
  trees for non-discriminative decision-making,'' in \emph{{AAAI}}.\hskip 1em
  plus 0.5em minus 0.4em\relax {AAAI} Press, 2019, pp. 1418--1426.

\bibitem{berk2017a}
R.~Berk, H.~Heidari, S.~Jabbari, M.~Joseph, M.~J. Kearns, J.~Morgenstern,
  S.~Neel, and A.~Roth, ``A convex framework for fair regression,''
  \emph{CoRR}, vol. abs/1706.02409, 2017.

\bibitem{pearl2009}
J.~Pearl, \emph{Causality: {{Models}}, {{Reasoning}} and {{Inference}}},
  2nd~ed.\hskip 1em plus 0.5em minus 0.4em\relax {Cambridge University Press},
  2009.

\bibitem{spirtes2016}
P.~Spirtes and K.~Zhang, ``Causal discovery and inference: Concepts and recent
  methodological advances,'' \emph{Applied Informatics}, vol.~3, no.~1, p.~3,
  2016.

\bibitem{glymour2019}
C.~Glymour, K.~Zhang, and P.~Spirtes, ``Review of {{Causal Discovery Methods
  Based}} on {{Graphical Models}},'' \emph{Frontiers in Genetics}, vol.~10,
  2019.

\bibitem{spirtes2013}
P.~Spirtes, C.~Meek, and T.~S. Richardson, ``Causal inference in the presence
  of latent variables and selection bias,'' \emph{CoRR}, vol. abs/1302.4983,
  2013.

\bibitem{spirtes1991}
P.~Spirtes and C.~Glymour, ``An {{Algorithm}} for {{Fast Recovery}} of {{Sparse
  Causal Graphs}},'' \emph{Social Science Computer Review}, vol.~9, no.~1, pp.
  62--72, 1991.

\bibitem{colombo2012}
D.~Colombo, M.~H. Maathuis, M.~Kalisch, and T.~S. Richardson, ``Learning
  high-dimensional directed acyclic graphs with latent and selection
  variables,'' \emph{The Annals of Statistics}, vol.~40, no.~1, pp. 294--321,
  2012.

\bibitem{vowels2021}
M.~J. Vowels, N.~C. Camg{\"{o}}z, and R.~Bowden, ``D'ya like dags? {A} survey
  on structure learning and causal discovery,'' \emph{{ACM} Comput. Surv.},
  vol.~55, no.~4, pp. 82:1--82:36, 2023.

\bibitem{kocaoglu2017}
M.~Kocaoglu, C.~Snyder, A.~G. Dimakis, and S.~Vishwanath, ``{CausalGAN}:
  Learning causal implicit generative models with adversarial training,'' in
  \emph{{ICLR} (Poster)}.\hskip 1em plus 0.5em minus 0.4em\relax
  OpenReview.net, 2018.

\bibitem{zhang2018b}
J.~Zhang and E.~Bareinboim, ``Fairness in decision-making - the causal
  explanation formula,'' in \emph{{AAAI}}.\hskip 1em plus 0.5em minus
  0.4em\relax {AAAI} Press, 2018, pp. 2037--2045.

\bibitem{kusner2018}
M.~J. Kusner, J.~R. Loftus, C.~Russell, and R.~Silva, ``Counterfactual
  fairness,'' in \emph{{NIPS}}, 2017, pp. 4066--4076.

\bibitem{gulrajani2017}
I.~Gulrajani, F.~Ahmed, M.~Arjovsky, V.~Dumoulin, and A.~C. Courville,
  ``Improved training of wasserstein {GANs},'' in \emph{{NIPS}}, 2017, pp.
  5767--5777.

\bibitem{kohavi1996}
R.~Kohavi, ``Scaling up the accuracy of naive-bayes classifiers: {A}
  decision-tree hybrid,'' in \emph{{KDD}}.\hskip 1em plus 0.5em minus
  0.4em\relax {AAAI} Press, 1996, pp. 202--207.

\bibitem{mattu}
Mattu, J.~Angwin, L.~Kirchner, Surya, and J.~Larson, ``How {{We Analyzed}} the
  {{COMPAS Recidivism Algorithm}},'' {ProPublica}, 2016,
  \href{https://www.propublica.org/article/how-we-analyzed-the-compas-recidivism-algorithm}{https://www.propublica.org/article/how-we-analyzed-the-compas-recidivism-algorithm}.

\bibitem{xu2018}
D.~Xu, S.~Yuan, L.~Zhang, and X.~Wu, ``Fairgan: Fairness-aware generative
  adversarial networks,'' in \emph{{IEEE} BigData}.\hskip 1em plus 0.5em minus
  0.4em\relax {IEEE}, 2018, pp. 570--575.

\bibitem{qureshi2020}
B.~Qureshi, F.~Kamiran, A.~Karim, S.~Ruggieri, and D.~Pedreschi, ``Causal
  inference for social discrimination reasoning,'' \emph{J. Intell. Inf.
  Syst.}, vol.~54, no.~2, pp. 425--437, 2020.

\bibitem{wu2008top}
X.~Wu, V.~Kumar, J.~R. Quinlan, J.~Ghosh, Q.~Yang, H.~Motoda, G.~J. McLachlan,
  A.~Ng, B.~Liu, S.~Y. Philip \emph{et~al.}, ``Top 10 algorithms in data
  mining,'' \emph{Knowledge and information systems}, vol.~14, no.~1, pp.
  1--37, 2008.

\bibitem{cox1958regression}
D.~R. Cox, ``The regression analysis of binary sequences,'' \emph{Journal of
  the Royal Statistical Society: Series B (Methodological)}, vol.~20, no.~2,
  pp. 215--232, 1958.

\bibitem{cortes1995support}
C.~Cortes and V.~Vapnik, ``Support-vector networks,'' \emph{Machine learning},
  vol.~20, no.~3, pp. 273--297, 1995.

\bibitem{ho1995random}
T.~K. Ho, ``The random subspace method for constructing decision forests,''
  \emph{{IEEE} Trans. Pattern Anal. Mach. Intell.}, vol.~20, no.~8, pp.
  832--844, 1998.

\bibitem{binkyte-sadauskiene2022}
R.~Binkyte{-}Sadauskiene, K.~Makhlouf, C.~Pinz{\'{o}}n, S.~Zhioua, and
  C.~Palamidessi, ``Causal discovery for fairness,'' \emph{CoRR}, vol.
  abs/2206.06685, 2022.

\bibitem{chae2018}
D.~Chae, J.~Kang, S.~Kim, and J.~Lee, ``{CFGAN:} {A} generic collaborative
  filtering framework based on generative adversarial networks,'' in
  \emph{{CIKM}}.\hskip 1em plus 0.5em minus 0.4em\relax {ACM}, 2018, pp.
  137--146.

\bibitem{plecko2021}
D.~Plecko, N.~Bennett, and N.~Meinshausen, ``fairadapt: Causal reasoning for
  fair data pre-processing,'' \emph{CoRR}, vol. abs/2110.10200, 2021.

\bibitem{NEURIPS2019_9015}
A.~Paszke \emph{et~al.}, ``Pytorch: An imperative style, high-performance deep
  learning library,'' in \emph{NeurIPS}, 2019, pp. 8024--8035.

\bibitem{diamond2016cvxpy}
S.~Diamond and S.~P. Boyd, ``{CVXPY:} {A} python-embedded modeling language for
  convex optimization,'' \emph{J. Mach. Learn. Res.}, vol.~17, pp. 83:1--83:5,
  2016.

\end{thebibliography}

\appendix

\section{Experiment Details and Results}

\subsection{Dataset and Training Details}\label{sec:dataset}

\begin{figure}
    \centering
    \includegraphics[width=80mm,height=1.5in]{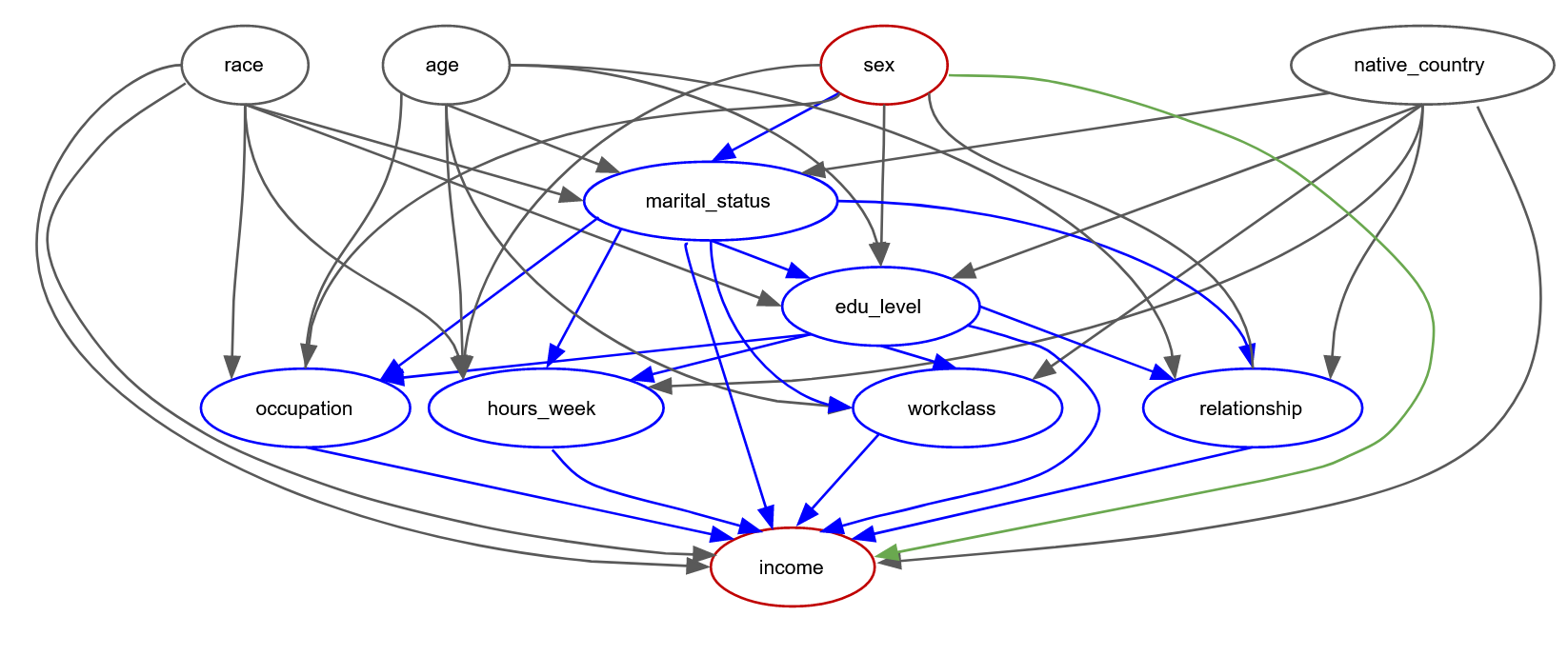}
    \caption{The causal graph of the Adult dataset depicts the indirect path set with blue paths, while the direct path is represented by the green path.}
    \label{fig:causal}
\end{figure}

\begin{figure}
    \centering
    \includegraphics[width=.49\textwidth]{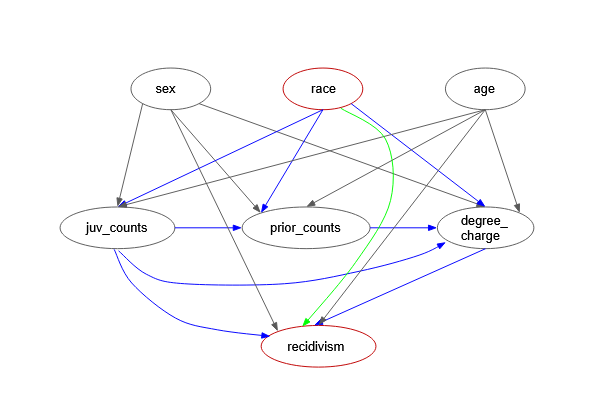}
    \caption{The causal graph of the COMPAS dataset depicts the indirect path set with blue paths, while the direct path is represented by the green path}
    \label{fig:causal_2}
\end{figure}

The causal graph \cite{chae2018} for ADULT is shown in Fig. \ref{fig:causal}, and for COMPAS \cite{plecko2021} in Fig. \ref{fig:causal_2}. Note that the causal graphs here are sourced from existing literature. 

\subsubsection{Adult Dataset}\label{adult} The Adult dataset was drawn from the 1994 United States Census Bureau data. It contains
65,123 samples with 11 variables. It used personal information such as education level and working hours per week to predict whether an individual earns more or less than \$50,000 per year. The dataset is imbalanced -- the instances made less than \$50,000 constitute 25\% of the dataset, and the instances made more than \$50,000 constitute 75\% of the dataset. As for gender, it is also imbalanced. We use age, years of education, capital gain, capital loss, hours-per-week, etc., as continuous features, and education level, gender, etc., as categorical features. We set the batch size at 640 and train 30 epochs for convergence. We set the learning rate $\eta$ at 0.001 according to the experiment result. 

\subsubsection{COMPAS Dataset}

COMPAS (Correctional Offender Management Profiling for Alternative Sanctions) is a popular commercial algorithm used by judges and parole officers for scoring criminal defendant’s likelihood of reoffending (recidivism). The COMPAS dataset includes the processed COMPAS data between 2013-2014. The data cleaning process followed the guidance in the original COMPAS repo. It Contains 6172 observations and 14 features. In our causal graph, we use 7 features. Due to the limited size of COMPAS dataset, it does not perform so well on NN based tasks.

\subsection{Training Details}\label{training}

For ADULT and COMPAS datasets, some pre-processing is performed. We normalize the continuous features and use one-hot encoding to deal with the categorical features for the input of $F^1$ and $F^2$. We use sex and race as the sensitive variable $S$ in ADULT and COMPAS respectively, income and two year recidivism as the outcome variable $Y$. 
%
% say explicitely the source of the graphs
%
 %The fairness threshold is 0.05, i.e., the effect should be in $\left [-0.05, 0.05\right ]$ to be fair. 
%We compare our method with other data generating approaches for different fairness respectively as other approaches may only be able to achieve one or two types of fairness.

For $F^1$ and $F^2$, we apply fully connected layers. For the discriminator $D$, we use the same architecture proposed in \cite{gulrajani2017}. We apply
SGD algorithm with a  momentum of 0.9 to update $F^1$ and $F^2$. $D$ is updated by the Adam algorithm with a learning rate 0.0001. Following \cite{gulrajani2017}, we adjust the learning rate $\eta$ by $\eta = \frac{0.01}{(1+10p)^{-0.75}}$ , where $p$ is the training progress linearly changing from 0
to 1. We update $F^1$ and $F^2$ 
for 2 steps then update $D$ for 1 step. For more details of the experiment (e.g., the split of training and testing datasets, the details of architectures of the neural nets, the estimation of Wasserstein distance), please refer to the Appendix (\ref{training}). We then evaluate the performance of our method of reweighting to achieve different types of causal fairness and utility. 

Our test are run on an Intel(r) Core(TM) i7-8700 CPU. The networks in the experiments are built based on Pytorch \cite{NEURIPS2019_9015},
the optimization in Equation \eqref{eq:w} is performed with the Python package CVXPY \cite{diamond2016cvxpy}. 

\subsubsection{Details of architectures of the feed-forward Neural Networks $F^1$ and $F^2$ with sub-neural networks} \label{feed_forward}
\begin{figure}
\centering     %%% not \center
\subfigure[Causal graph $\mathcal{G}$ ]{\label{neural_1}\includegraphics[width=60mm,height=1.9in]{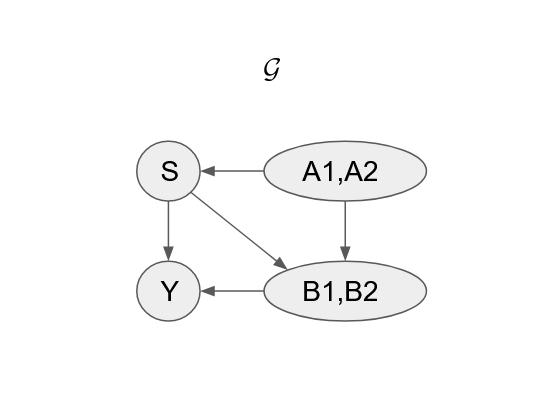}}
\subfigure[Neural Network $F^1$]{\label{neural_2}\includegraphics[width=80mm,height=1.5in]{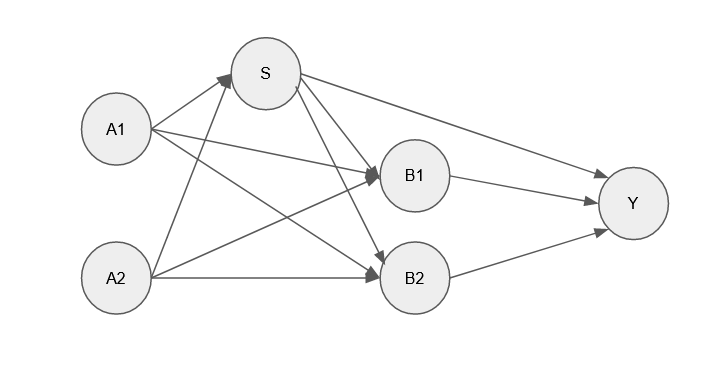}}
\caption{details of the connection of the neural nets of a given $\mathcal{G}$. In Fig. \ref{neural_2}, each nodes are either input or output of a sub-neural nets or both. Note that we do not show the inner layers here for simplicity. 
} \label{neural_nets}
\end{figure}

To simplify our demonstration, we consider a causal graph $\mathcal{G}$ with 6 attributes $\{S,A_1,A_2,B_1,B_2,Y\}$ as shown in Fig. \ref{neural_1}. And Fig. \ref{neural_2} shows the joint neural network of it. 

\subsubsection{Details of WGAN-GP adaptation for our method} In our design, we adopt the discriminator from WGAN-GP: in the original work, the discriminator is used to differentiate between the generated and real data while we are trying to differentiate between $S^+$ and $S^-$. The difference between orginal GAN and WGAN-GP is that WGAN-GP introduces a gradient penalty term in the training objective to guarantee Wasserstein distance. Wasserstein distance itself has been used a lot in fairness realted topic to help detect or mitigate bias. Note that we choose relatively larger batch size since to approximate Wasserstein distance between two distributions requires relatively larger batch size.

\begin{figure}
\centering     %%% not \center
\subfigure[Causal graph $\mathcal{G}$ and interventional graph $\mathcal{G}_s$ with the indirect interventional path $\pi_C$]{\label{fig:indirect_a}\includegraphics[width=50mm,height=1.2in]{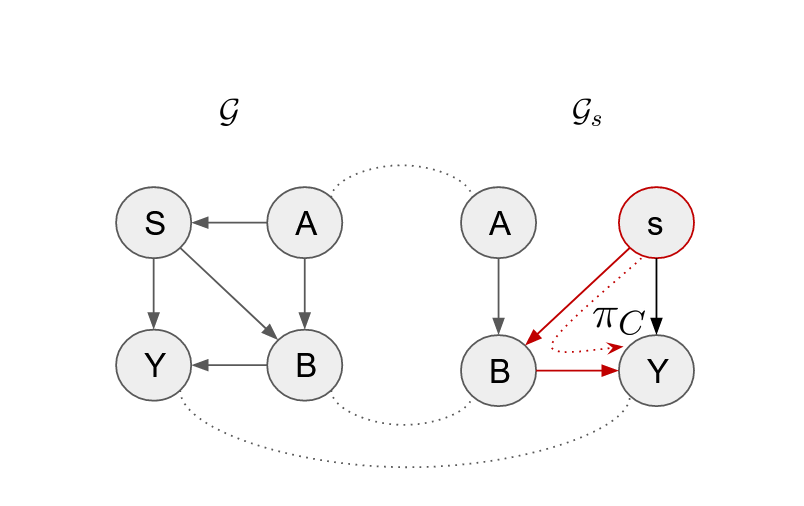}}
\subfigure[Neural Networks $F^1$ and $F^2$]{\label{fig:indirect_b}\includegraphics[width=70mm,height=1.5in]{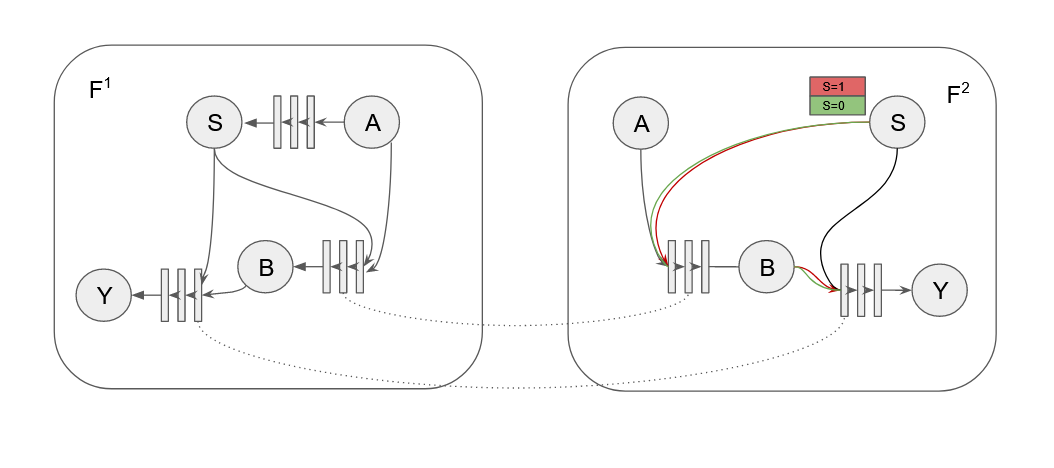}}
\caption{the Neural Networks $F^1$
and $F^2$ based on indrect discrimination. $S$ is 1 or 0 and the intervention
is only along $\pi_C= \{S\shortrightarrow B\shortrightarrow  Y \}$ for the
interventional distributions $P_{F^2} (s^+)$ (red) and $P_{F^2} (s^-)$  (green) respectively. Compared with Fig. \ref{fig:graph2}, we could see that the intervention is not transferred directly from $S$ to $Y$ ($\{S\shortrightarrow Y \}$) in Fig. \ref{fig:graph3}. 
} \label{fig:graph3}
\end{figure}

\subsubsection{Sensitivity to the Choice of Hyper-Parameters}
%We have also analyzed the sensitivity of our method to the hyper-parameters mentioned in Section \ref{method}, in Figure \ref{fig:t}. Figures indicates that the performance of our adversarial reweighting scheme has low sensitivity to the choice of the hyper-parameters once $T$ is higher than 1. So we set $T$ at 1. For analysis of other datasets, we refer the readers to Appendix \ref{training}. %\ref{fig:sens2} in Appendix %(\ref{sec:training-tabular}).

\begin{figure}
    \centering
    \includegraphics[width=.3\textwidth]{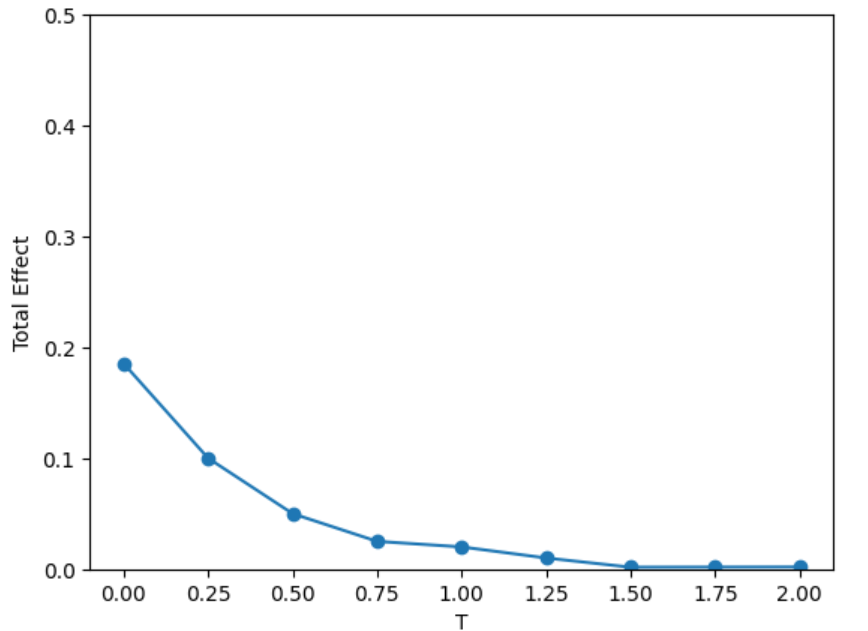}
    \caption{Sensitivity of total effect on the change of $T$ on ADULT dataset.}
    \label{fig:t}
\end{figure}

We conduct an analysis of the sensitivity of our method to the hyper-parameters discussed in Section \ref{method}, and the results are shown in Fig. \ref{fig:t}. The figures demonstrate that our adversarial reweighting scheme's performance has low sensitivity to hyper-parameter choice when $T$ is above 1. Therefore, we set $T$ at 1.5. 

\begin{figure}
    \centering
    \includegraphics[width=.3\textwidth]{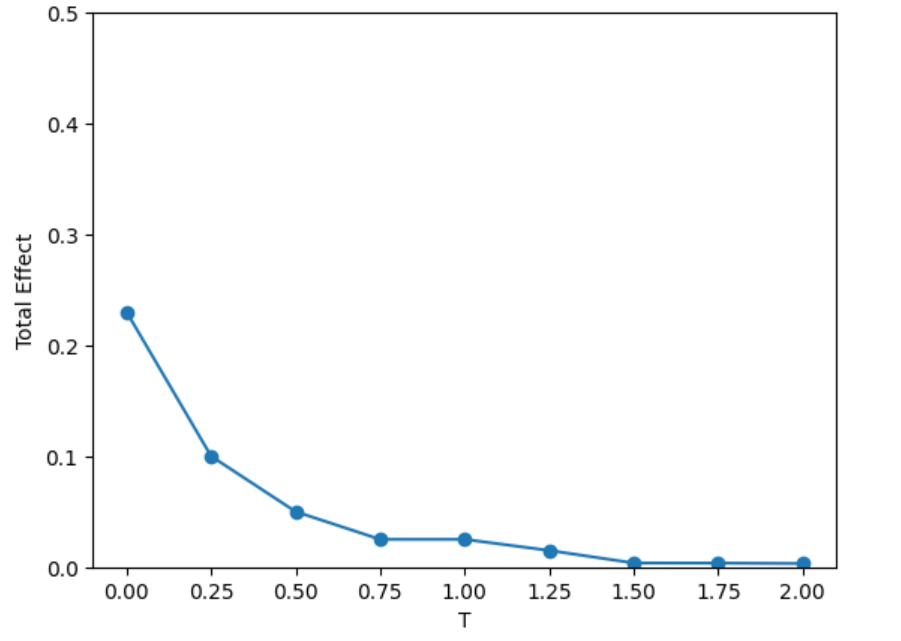}
    \caption{Sensitivity of total effect on the change of $T$ on COMPAS dataset.}
    \label{fig:t_2}
\end{figure}

We could see from the Fig. \ref{fig:t_2} that the trend of $T$ on total effect on COMPAS dataset is similar to what is shown earlier on ADULT dataset.
\subsubsection{Details of calculating different causal effects}
We have a discriminator in our design, to calculate different causal effect, we just send the samples from different groups ($S^+$ and $S^-$), normalize the output of $D$ within groups, then get the difference as causal effect. 

\subsubsection{Repetition} We repeat experiments on each dataset five times. Before each repetition, we randomly split data into training data (80\%) and test data (20\%) for the computation of the standard errors of the metrics. We train 30 epochs for convergence.

\subsubsection{Choices of Counterfactual Effect}\label{counterfactual}

Due to space limit, we only show two combinations of the counterfactual effect on individual features. For ADULT, we use $o_1=\{white,us\}$, $o_2=\{non\_white,us\}$. For COMPAS, we use $o_1=\{male,under\_25\}$, $o_2=\{male,above\_25\}$

\subsubsection{Details to approximate Wasserstein distance}\label{w_reweighting} To approximate the Wasserstein distance, we WGAN-GP discriminator between the orginal data and data for evaluation. When the NNs are trained, we use the discriminator to approximate the Wasserstein distance between the two datasets.

\end{document}